\documentclass{article} 
\usepackage{colm2024_conference}
\colmfinalcopy

\usepackage{microtype}
\usepackage{hyperref}
\usepackage{url}
\usepackage{booktabs}
\definecolor{darkblue}{rgb}{0, 0, 0.5}

\usepackage{amsmath}
\usepackage{amssymb}
\usepackage[utf8]{inputenc}
\usepackage{wrapfig}
\usepackage{setspace}
\usepackage{hyperref}
\usepackage{hhline}
\usepackage{booktabs}
\usepackage{latexsym}
\usepackage{soul}
\usepackage{stfloats}
\usepackage[most]{tcolorbox}
\usepackage{stmaryrd}
\usepackage{amsthm}
\usepackage{thmtools}
\usepackage{footnote}
\makesavenoteenv{tabular}
\definecolor{White}{rgb}{1, 1, 1}
\definecolor{Periwinkle}{rgb}{0, 0, 0}
\definecolor{myblue}{rgb}{0.82, 0.94, 0.75}
\definecolor{mygreen}{rgb}{0.64, 0.76, 0.68}
\definecolor{myyellow}{rgb}{0.88, 0.54, 0.35}
\definecolor{mygreen}{rgb}{0.68, 0.9, 0.8}
\definecolor{mypink}{rgb}{0.2, 0.87, 0.2}
\usepackage{booktabs} 

\newcommand{\GAP}{\textsc{DG-Diff}}

\colorlet{LightGray}{White!98!Periwinkle}

\declaretheoremstyle[
    name=\textsc{Self-[In]Correct},
]{thmsty}
\declaretheorem[style=thmsty,numbered=no]{hypothesis}
\tcolorboxenvironment{hypothesis}{enhanced jigsaw,colback=LightGray,drop shadow,boxrule=0.9pt, boxsep=0.1pt,left=4pt,right=4pt,top=4pt,bottom=4pt}

\declaretheoremstyle[
    name=\textsc{Sub H1},
]{thmsty2}

\tcolorboxenvironment{subhypothesisone}{enhanced jigsaw,colback=LightGray,drop shadow,boxrule=0.9pt, boxsep=0.1pt,left=4pt,right=4pt,top=4pt,bottom=4pt}

\declaretheoremstyle[
    name=\textsc{Sub H2},
]{thmsty3}

\tcolorboxenvironment{subhypothesistwo}{enhanced jigsaw,colback=LightGray,drop shadow,boxrule=0.9pt, boxsep=0.1pt,left=4pt,right=4pt,top=4pt,bottom=4pt}

\usepackage[utf8]{inputenc}

\usepackage{microtype}
\usepackage{tabularx}
\usepackage{multirow}
\usepackage{booktabs}
\usepackage{graphicx}
\usepackage{listings}
\usepackage{color}
\usepackage{xcolor}
\usepackage[utf8]{inputenc}
\usepackage{listings}
\usepackage{amssymb}
\usepackage{enumitem}
\usepackage{adjustbox}
\usepackage{comment}
\usepackage{caption}
\usepackage{cleveref}
\crefformat{section}{\textcolor{blue}{\S#2#1#3}} 
\crefformat{subsection}{\textcolor{blue}{\S#2#1#3}}
\crefformat{subsubsection}{\textcolor{blue}{\S#2#1#3}}
\crefformat{equation}{\textcolor{orange}{Eq.~(#1)}}

\usepackage{color}

\definecolor{darkred}{rgb}{0.6,0.0,0.0}
\definecolor{darkgreen}{rgb}{0,0.50,0}
\definecolor{lightblue}{rgb}{0.0,0.42,0.91}
\definecolor{orange}{rgb}{0.99,0.48,0.13}
\definecolor{grass}{rgb}{0.18,0.80,0.18}
\definecolor{pink}{rgb}{0.97,0.15,0.45}
\definecolor{codegreen}{rgb}{0,0.6,0}
\definecolor{codegray}{rgb}{0.5,0.5,0.5}
\definecolor{codepurple}{rgb}{0.58,0,0.82}
\definecolor{backcolour}{rgb}{0.95,0.95,0.92}
\definecolor{keywordcolor}{rgb}{0.7, 0.1, 0.1}   
\definecolor{tacticcolor}{rgb}{0.0, 0.1, 0.6}    
\definecolor{commentcolor}{rgb}{0.4, 0.4, 0.4}   
\definecolor{symbolcolor}{rgb}{0.0, 0.1, 0.6}    
\definecolor{sortcolor}{rgb}{0.1, 0.5, 0.1}      
\definecolor{attributecolor}{rgb}{0.7, 0.1, 0.1} 

\usepackage{float}
\DeclareUnicodeCharacter{2200}{\forall}
\DeclareUnicodeCharacter{2227}{\logicand}
\DeclareUnicodeCharacter{2192}{\rightarrow}

\def\arrvline{\hfil\kern\arraycolsep\vline\kern-\arraycolsep\hfilneg}
\newcommand{\PHENOMENON}{\textsc{Self-[In]Correct}}
\newcommand{\hyperPHENOMENON}{\hyperlink{main hypothesis}{\textsc{Self-[In]Correct}}}

\title{\PHENOMENON{}:  LLMs Struggle with Discriminating Self-Generated Responses}

\author{Antiquus S.~Hippocampus, Natalia Cerebro \& Amelie P. Amygdale \thanks{ Use footnote for providing further information
about author (webpage, alternative address)---\emph{not} for acknowledging
funding agencies.  Funding acknowledgements go at the end of the paper.} \\
Department of Computer Science\\
Cranberry-Lemon University\\
Pittsburgh, PA 15213, USA \\
\texttt{\{hippo,brain,jen\}@cs.cranberry-lemon.edu} \\
\And
Ji Q. Ren \& Yevgeny LeNet \\
Department of Computational Neuroscience \\
University of the Witwatersrand \\
Joburg, South Africa \\
\texttt{\{robot,net\}@wits.ac.za} \\
\AND
Coauthor \\
Affiliation \\
Address \\
\texttt{email}
}

\newcommand{\theoremsymbol}{$\textbf{H}_0$}

\author{Dongwei Jiang \quad Jingyu Zhang \quad Orion Weller  \quad  \textbf{Nathaniel Weir} \\ 
\textbf{Benjamin Van Durme} \quad \textbf{Daniel Khashabi} \quad \\
   Johns Hopkins University \\
  \texttt{\{djiang21, jzhan237\}@jhu.edu}
}

\begin{document}

\maketitle

\begin{abstract}
Can LLMs consistently improve their previous outputs for better results?
For this to be true, LLMs would need 
to be better at \emph{discriminating} among previously-generated alternatives, than \emph{generating} initial responses.  
We explore the validity of this hypothesis in practice.
We first formulate a unified framework that allows us to compare the generative and discriminative capability of any model on any task.
In our resulting experimental analysis of several open-source and industrial LLMs, 
we observe that models are not reliably better at discriminating among previously-generated alternatives than generating initial responses.
This finding challenges the notion that LLMs may be able to enhance their performance \emph{only} through their \emph{own} 
judgment. 
\end{abstract}

\section{Introduction}
The promise of Large Language Models (LLMs) that can self-improve has brought both excitement and fear about the future impact of AI. 
However, it remains a mystery what is needed for LLMs to continually self-improve \citep{cant}.\footnote{
\label{footnote:1}
Our study focuses on scenarios where LLMs utilize their \underline{\emph{own}} 
judgement (hence the term ``self-...''). This is fundamentally different from scenarios involving \emph{external} feedback, as examined in prior research~\citep{wang2023voyager}. }
A crucial aspect of human learning involves reflecting on one's actions. This self-improvement is feasible because individuals can identify their own mistakes and adjust their future decisions accordingly \citep{mayo1996error, corder1967significance}.
This principle should be applicable to LLMs as well. 

For LLMs to reliably self-improve 
based on their decisions, 
the ability to \emph{discriminate} (distinguish) the goodness of their own prior generations
 should surpass the ability to \emph{generate} good solutions directly. 
Given the importance of this capability, it is worth raising a question about the foundations of self-discrimination: 
\emph{Are LLMs really better at discrimination than generation?}

\begin{figure*}[h!]
    \centering

    \includegraphics[trim=0.0cm 7.5cm 0cm 0.2cm,scale=0.54]{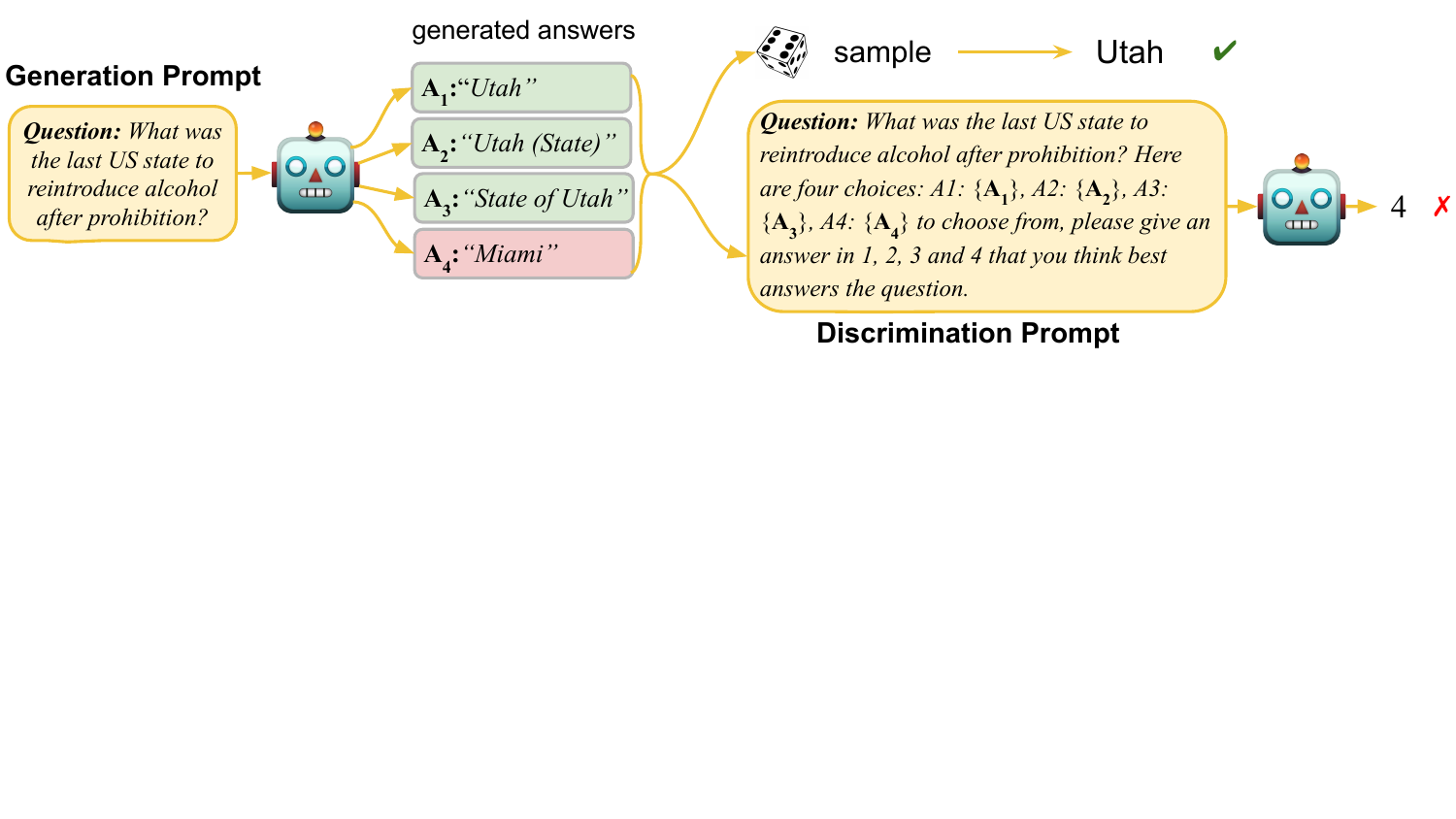}
    \captionsetup{belowskip=-10pt} 
    \caption{\textbf{Two phases evaluated in our paper.} In the generation phase, the model is fed with generation prompt. Generated answers are collected and randomly selected to calculate the generation score. In the discrimination phase, the model is fed with discrimination prompt and generated answers. The model would choose between generated answers and the score of the chosen answer would be used for calculating the discrimination score}
    \label{fig:teaser}
\end{figure*}


This paper seeks to answer this question by proposing the \hyperPHENOMENON{} hypothesis (\cref{main_hypothesis}):  \textit{LLMs are not reliably better at discriminating among previously-generated alternatives than generating initial responses}. 
Determining the validity of this hypothesis is crucial, as existing studies provide initial evidence suggesting that the capability to distinguish between LLM-generated options is both a sufficient \cite{can_correct} and necessary \cite{cant} condition for self-improvement.

It is non-trivial to compare LLMs' generative capability with their discriminative capability on the same footing. 
\citet{paradox} compares these two capabilities, albeit in a slightly different setting. \citet{paradox} measure a model's ability to discriminate (identify) the \textit{ground-truth answer} among distractor options. However, ground-truth answers are not always likely to be among a model's generated outputs. In contrast, our work quantify a model's ability to discriminate its \emph{self-generated} candidate answers, which is more aligned with the mechanisms of LLM self-improvement.
To better measure these abilities, we implement a two-phase methodology depicted in \autoref{fig:teaser}. In the first phase, we generate multiple outputs using a temperature setting greater than zero, then randomly select one of these outputs, using its evaluation as an indicator of generative performance. In the second phase, we instruct the LLM to choose the best answer from its own outputs, with the evaluation of the selected answer serving as the indicator of discriminative performance. This approach is consistent with the actual procedure employed in various self-improvement studies~\cite{self-refine, self-rewarding}. 
Further details can be found in \S\ref{sec:hypothesis_formulation}.

To back up the \PHENOMENON{} hypothesis,
we conduct experiments covering widely used LLMs (Phi-3, LLaMA series, Mixtral series, and GPT series) on a diverse set of tasks including mathematics, world knowledge acquisition, truthful question answering, and instructions following. 
Our investigation in \S\ref{sec:Main Results} reveals a surprising finding: 
while there is evidence that humans find the task of discrimination simpler than generation~\cite{expertise1}, 
on various evaluated tasks we observed that 
LLMs are not consistently better at discriminating among previously-generated alternatives than generating
initial responses.

We also conduct a series of analyses to deepen our understanding of \PHENOMENON. First, we investigate alternative prompt choices (\S\ref{sec:additional}) to ensure that \PHENOMENON{} is not simply a result of sub-optimal prompt design. Our findings indicate that \PHENOMENON{} persists with the inclusion of additional in-context learning examples and chain-of-thought demonstrations. Next, we examine the role of pre-training objectives (\S\ref{sec:pre-training objective}) and discover that \PHENOMENON{} does not appear in \textit{non}-autoregressive models (e.g., the FLAN family). Third, we discuss the apparent contradiction of our method with recent studies on self-improvement \citep{self-refine, self-rewarding} (\S\ref{sec:self-reflection}). Finally, we highlight the potential implications of our findings in \S\ref{sec:implication}.


Our contributions in this paper are two-fold:
\begin{itemize}[noitemsep,topsep=0pt,parsep=0pt,partopsep=0pt,leftmargin=10pt]
    \item We develop a unified framework that facilitates the testing of both generative and discriminative capabilities of any LLM on any task; 
    \item We conduct experiments on widely used LLMs and collected empirical evidence to support \PHENOMENON{}. We also provided additional experiments to better understand \PHENOMENON{} and its implications.
\end{itemize}

\section{Related Work}
\paragraph{Self-Improvement with LLMs.} The concept of self-improvement existed before the LLM era. Earlier approaches employed generative adversarial networks (GANs)~\citep{DBLP:conf/rep4nlp/SubramanianRDPC17, DBLP:conf/aaai/YuZWY17} to improve NLP systems via self-generated feedback. \citet{welleck2022selfcorrect} trained a separate corrector model to iteratively refine generations.

In the era of LLM, self-improvement with self-feedback has also been studied in various forms~\citep{Pan2023AutomaticallyCL, saunders2022selfcritiquingmodelsassistinghuman}.
Self-Instruct~\citep{wang2023selfinstruct} improves the instruction-following capabilities of pre-trained language models by bootstrapping off their own generations.
\citet{self-rewarding} employs LLMs to provide rewards for their own generation. 
\citet{self-play} uses a self-play mechanism where the LLM refines its capability by playing against instances of itself. \citet{Reflexion} achieves self-improvement by having the model generate verbal reflection on its own outputs at inference time.
Several other recent studies \citep{self-refine, liu2023crystal, butt2024codeit, krishna2023intersectionselfcorrectiontrustlanguage, DBLP:journals/corr/abs-2212-10560} also adopted this idea and applied it to different tasks.

The success stories mentioned in previous paragraph show that when
  external ground-truth feedback is available, LLMs can effectively engage in self-improvement. \citet{gou2024critic} shows that LLMs can verify and correct their initial responses through interactions with various external tools. Similarly, \citet{can_correct} and \citet{Reflexion} have shown that ground-truth feedback can significantly enhance LLM performance across various tasks.


However, in the absence of ground-truth feedback—where LLMs must refine their initial answers based solely on their inherent capabilities (i.e., intrinsic self-improvement)—the situation changes. Critics have argued~\citep{cant, valmeekam2023large, can_correct} that the reported self-improvement on reasoning tasks may be no more effective than self-consistency~\citep{wang2022self}, and that such improvements are often a result of an inferior initial response. Our work adopts a similar intrinsic self-improvement setting, and we explore LLMs' generative and discriminative capacities beyond reasoning tasks.

\paragraph{Discrepancy between LLM generation and discrimination.}

For humans, distinguishing a good solution from a bad one is often easier than coming up with a solution from scratch~\citep{expertise1}. However, recent studies are starting to question if the same applies to LLMs. \citet{paradox} and \citet{tan2024iunderstandicreate} investigated multiple NLP tasks, and showed that LLMs often struggle to understand their own outputs. 
To evaluate the generation and discrimination performance of LLMs, \citet{DBLP:journals/corr/abs-2311-09184} conducted experiments focusing on summarization.
\citet{arora2023learning} and \citet{chen2024treesearchusefulllm} conducted similar experiments in the domain of planning. 
Our work differs from previous research by evaluating this discrepancy on a wider range of tasks using a unified metric while also trying to uncover the reasons behind it.

\paragraph{Using LLMs for self-evaluation. }
Recent studies indicate the potential of LLM evaluation that is close to human level~\citep{DBLP:conf/acl/ChiangL23, Gilardi_2023, lin2024criticbenchbenchmarkingllmscritiquecorrect}. However, for the task of self-evaluation, concerns have been raised by~\citet{valmeekam2023large} and \citet{cant}, who pointed out that LLM encounters difficulties in self-evaluating its generation for math tasks. Further research by~\citet{stechly2024selfverification}, \citet{gpt4doesntknow} and \citet{valmeekam2023large} has uncovered models' limitations on self-evaluation for tasks requiring complex reasoning and planning. 
Compared to these works, our work seeks to explore the efficacy of LLM self-evaluation in a broader range of tasks.

\section{\PHENOMENON}
\label{section:method}

We formally define our evaluation setting 
( \autoref{fig:teaser}), 
and present our hypothesis.

\subsection{Establishing an Evaluation Criteria to Compare Generation vs. Discrimination} 
\label{sec:hypothesis}

Given a task $T$ with an evaluation dataset \(D = \{(x_i, y_i)\}_{i=1}^{m}\) and evaluation metric $f$, we use the same LLM, denoted by $P_\text{LM}$, for both generation and discrimination. For each evaluation input $x_i$, we first sample $n$ candidate generations $g_1(x_i),\dots,g_n(x_i) \sim P_\text{LM}(x_i)$ using the default task prompt (generation prompt). We use a low temperature during sampling to ensure the generated outputs are all highly probable.

\paragraph{Evaluating generation.} 
The performance of the \textit{generative} phase for each evaluation sample \(x_i\) is computed 
applying the evaluation metric $f$ to 
a randomly chosen generation from the \(n\) candidate generations \(G(
x_i)=\{ g_1(x_i), g_2(x_i), \ldots, g_n(x_i) \}\): 
\begin{equation*}
S_{\text{gen}}(x_i) = f\Bigl(g_\text{rand}(x_i), y_i\Bigr),
\end{equation*}
where \(g_\text{rand}\) is a randomly-sampled generation \(g_\text{rand}(x_i)\sim G(x_i) \).
The notation \(f(g_j(x_i), y_i)\) represents the metric \(f\) applied to the \(j\)-th generation output for the \(i\)-th evaluation sample and  \(g_\text{rand}(x_i)\) represents one random candidate generations for that sample. Because the candidate generations for each sample are produced by sampling from the language model \(P_{\text{LM}}\) 
using the same hyper-parameters (temperature, top-$p$, etc), choosing a random candidate from $G(x_i)$ is essentially equivalent to generating an output directly from \(P_{\text{LM}}(x_i)\). 
The overall generation performance \(S_{\text{gen}}\) is the average of  \(S_\text{gen}(x_i)\) across all samples: 
$$S_{\text{gen}} = \frac{1}{m}\sum_{i=1}^m S_\text{gen}(x_i).$$ 

\paragraph{Evaluating discrimination.} To assess the discrimination performance, we feed the generations back to \(P_{\text{LM}}\) and prompt it to identify the most suitable answer. For each task \(T\), we construct a discriminative prompt 
\(p_{\text{disc},T}\) 
(the prompts are available in \autoref{app: COT}) and feed it the $n$-many generated responses.
Note that the ordering of generated candidates is always random as the sampling is conducted uniformly with temperature $> 0$. We tried reordering the candidates before sending them for discrimination and the result remains very similar.
Using few-shot prompting, we guide \(P_{\text{LM}}\) to output the label of the preferred \texttt{chosen} answer and the label \texttt{chosen} in \(\{1, 2, \ldots, n\}\) is determined by greedily decoding the output of \(P_{\text{LM}}(\cdot|x_{\text{disc},T}(G(x_i)))\). The discrimination performance for each sample \(i\) is quantified by:
\begin{equation*}
S_\text{disc}(x_i) = f\Big(g_\text{chosen}(x_i), y_i \Big).
\end{equation*}
To derive an overall measure of discrimination performance \(S_{\text{disc}}\), we average the individual scores \(S_\text{disc}(x_i)\) across all samples:
\vspace{-0.2cm}
$$S_{\text{disc}} = \frac{1}{m}\sum_{i=1}^m S_\text{disc}(x_i).$$



We also consider evaluating discriminative ability by calculating the absolute score of each candidate separately and selecting the best candidate. But we didn't find much difference compared to the setup here (details in \autoref{app:evaluate_separately}).

\subsection{Hypothesis Formulation} \label{sec:hypothesis_formulation}
Given the above definitions, our main hypothesis becomes easy to formalize. For any given task, denote \GAP{} as the \textbf{difference between discrimination performance and generation performance},
\begin{equation*}
    \GAP = S_{\text{disc}} - S_{\text{gen}}. \label{DG-DIFF}
\end{equation*}

Our \hypertarget{main hypothesis}{main hypothesis} is:
\vspace{-0.1cm}
    \begin{hypothesis}
        LLMs are not reliably better at discriminating among previously generated alternatives ($S_\text{disc}$) than generating initial responses ($S_\text{gen}$)
        and hence,    
         $\GAP = S_\text{disc} - S_\text{gen} \leq 0$. 
    \end{hypothesis}
            \label{main_hypothesis}
\vspace{0.0cm}

\paragraph{Hypothesis testing for \PHENOMENON.} 
To validate \PHENOMENON{}, one can apply the framework of statistical hypothesis testing~\cite{dror2018hitchhiker,sadeqi-azer2020not}. We treat \PHENOMENON{} as the null hypothesis (\theoremsymbol) to provide an objective basis for testing.
In this context, the conventional wisdom that discrimination is better than generation serves as the alternative hypothesis ($\textbf{H}_1$). To reject the null hypothesis \theoremsymbol, it must be demonstrated that $\GAP$ is a sufficiently large positive value to justify its rejection.
Details of the hypothesis testing on our experimental datasets are provided in \S\ref{sec: experimental_setup}.

\paragraph{Design choices for hypothesis testing.} An important design choice in our framework is that the candidate generations $G(x_i)$ are shared across the generative and discriminative phases. This design choice allows us to formulate the generative phase as a \textit{random multiple choice} among pre-generated candidates. As a result, it allows a fair comparison with the discriminative phase, where the task is \textit{using LLM for multiple choice} among the same candidates.\footnote{Under this setup, the generative and discriminative phases will always have the same upper/lower bound of possible scores.}

Our framework applies the task's original metrics in both the generative and discriminative phases, which ensures consistency across assessments. By eliminating the need for human input, our framework is more scalable and cost-effective than ~\citet{paradox} and \citet{llm-as-a-judge}, which depend on human annotation for discrimination. Our metrics are also closely aligned to the actual process that's employed in self-improvement literature \citep{Reflexion, self-refine}, where the model is asked to choose the best answer from a list of generations. Nevertheless, we would like to mention that because generation and discrimination are two very different processes, the metrics used in this paper are only proxies to evaluate those two important capabilities.

\section{Empirical Support for \PHENOMENON}


In this section, we describe our experimental setup (\S\ref{sec: experimental_setup}) and lay out the main findings (\S\ref{sec:Main Results}).

\makesavenoteenv{tabular}
\begin{table*}
    \centering
    \small
    \begin{tabular}{lcccccc}
        \toprule
        \textbf{Task}& \textbf{Split} & \textbf{\#Eval} & \textbf{\#Shots} &   \textbf{Task Type} & \textbf{Metric $f(.)$} & \textbf{Metric Type} \\
        \midrule
        \textbf{GSM8K} & Test & 1319 & 2  & Math Word Problem  & Accuracy & Binary  \\
        \textbf{TriviaQA} & Val & 17944 & 2  & Question Answering & Accuracy & Binary \\
        \textbf{MT-Bench} & Test & 160 & 3 & Instruction Following  & GPT-4 score & Categorical \\
        \textbf{TruthfulQA} & Val  & 817 & 2 & Question Answering & GPT-judge & Binary  \\
        \bottomrule
    \end{tabular}
    \caption{Configuration of experimental tasks. ``Split'' specifies which subset the data originates from. ``\#Eval'' indicates the number of instances used for evaluation. ``\#Shots'' specifies the number of few-shot examples employed for evaluation. To evaluate TruthfulQA generations, we follow \citet{lin-etal-2022-truthfulqa} and develop two ``GPT-judges'' by fine-tuning GPT-3 models with provided data.
    }
    
    \label{tab:setup}
\end{table*}
\makesavenoteenv{table}

\subsection{Experimental Setup} \label{sec: experimental_setup}
\paragraph{Tasks.} 
A summary of the tasks we evaluate on is provided in \autoref{tab:setup}.
We assess our hypothesis on a diverse set of tasks including GSM8K~\citep{gsm8k} for math, TriviaQA~\citep{TriviaQA} for world knowledge, TruthfulQA~\citep{lin-etal-2022-truthfulqa} for truthfulness in question answering, and MT-Bench \citep{llm-as-a-judge} for instruction following. These represent a diverse set of benchmarks used to evaluate 
LLMs across various domains. 
For TriviaQA, we use the \texttt{rc.nocontext} setup, which means the model relies solely on its parametric knowledge to answer the question correctly without accompanying context or documents. For TruthfulQA, we use the generation setup, where the model generates responses to a set of questions. The metrics scale for MT-Bench is 0-10 \citep{llm-as-a-judge}.

\paragraph{Task metrics.} 
The list of task-specific metrics $f(.)$ is provided in  \autoref{tab:setup}. 
The evaluation for GSM8K, TriviaQA and TruthfulQA is conducted using \texttt{lm-evaluation-harness}\footnote{\url{https://github.com/EleutherAI/lm-evaluation-harness}}, which provides a standardized framework for assessing model performance across benchmarks. The evaluation for MT-Bench is done with \texttt{llm\_judge}\footnote{\url{https://github.com/lm-sys/FastChat/tree/main/fastchat/llm_judge}}, which use GPT-4 score \citep{llm-as-a-judge} to score the generated answer from models. We do not test GPT-4-turbo on MT-Bench to avoid self-evaluation bias \citep{he-etal-2023-blind}. To evaluate TruthfulQA, we follow \citet{lin-etal-2022-truthfulqa} and develop two ``GPT-judges'' by fine-tuning GPT-3 models\footnote{The original ``GPT-judges'' were fine-tuned with \texttt{curie} models which are no longer available for fine-tuning. Therefore, we use \texttt{davinci-002} which is larger than \texttt{curie}.} with provided data
. Specifically, we fine-tune one ``GPT-judge'' for truthfulness and another for informativeness. 
Finally, we report the percentage of answers that are both truthful and informative as the final metric for TruthfulQA. 

\paragraph{Hypothesis Testing for \PHENOMENON{} Across Tasks.} We apply a one-sided McNemar's Test \citep{Mcnemar1947NoteOT} for GSM8K, TriviaQA, and TruthfulQA to calculate p-values and assess statistical significance, as this test is ideal for binary outcome comparisons. For MT-Bench, we use the Wilcoxon signed-rank test \citep{c4091bd3-d888-3152-8886-c284bf66a93a} because it handles categorical data and does not assume a normal distribution. Further details on our test selection and hypothesis testing methodology are provided in \autoref{app:statistial_test}.

\paragraph{Handling failure modes during evaluation.}
During the evaluation of the discrimination phase, if the model's output does not conform to the expected format (i.e., integers indicating the selected answer), we consider it a failure. While such an output would receive a score of 0 in the generative setting, we take a more lenient approach for the discrimination phase. In these cases, we assign the model the score of the lowest-performing generated answer (according to our metric $f$) among the other candidate answers: $S_\text{disc}(x_i) = \min_{g(x_i) \in G(x_i)} f(g(x_i), y_i)$. We also try to make the discriminator output one of the answers directly in the case of a failure, hoping that bypassing this extra step of identifying the multiple-choice options would simplify the discrimination phase. However, we observe an increased percentage of invalid discrimination output with similar discrimination performance (see \autoref{app:additional_experiments} for more details).  In our experiments, we observe that the average rate of invalid responses remained low (often less than 5\%). Given that there is also a small proportion of invalid outputs in the generation phase that wouldn't get any credit when selected, we believe that the occurrence of invalid discrimination outputs does not significantly impact our overall findings.

\paragraph{Models.} We employ a range of models 
including Phi-3-mini-4k-Instruct \citep{abdin2024phi3technicalreporthighly}, LLaMA-2 Base models (7B, 13B, and 70B), LLaMA-2 Chat models (7B, 13B and 70B), LLaMa-3 Base models (8B and 70B), LLaMa-3 Instruct models (8B and 70B), Mixtral \(8\times7\)B-Instruct-v0.1, GPT-3.5-turbo and GPT-4-turbo.\footnote{We use GPT-3.5-turbo-0125 and GPT-4-0125-preview. } 
For the evaluation of each model, we adapt our prompts to be compatible with the keywords used in their [pre-]training. For example, when prompting LLaMA-2 Chat models we use \texttt{<SYS>}, \texttt{<INST>} keywords to indicate system and instruction prompts. 

\paragraph{Model hyper-parameters.} During the generation phase, we use the default hyperparameter specified in \texttt{lm-eval-harness} for all tasks,
except for temperature, which we have adjusted to $0.7$.
We use an above $0$ temperature to obtain distinct generations upon multiple rounds of sampling. At the same time, during the discrimination phase,
we set the temperature to 0 to avoid any randomness.

\subsection{Main Findings}\label{sec:Main Results}


\paragraph{On a dominant majority of experiments, \PHENOMENON{} is \underline{not} rejected.}
Based on the results in \autoref{tab:main_result}, in 54 out of 56 experiments, the p-value exceeded the significance level (0.05), leading to the failure to reject the \PHENOMENON{} hypothesis. 
In fact, \GAP{} is generally small or negative
across both pre-trained models and aligned models.  
To test the effect of prompt variations, we conduct an ablation experiment in \autoref{app:ablation_of_prompts} and find that these variations do not significantly affect \GAP{}.
Although 
in few cases (2 out of 56) the p-value is high enough to reject our hypothesis (p-value $>0.05$), such as LLaMA-2-70B and GPT-3.5-turbo on TriviaQA, \GAP{} remains quite small in such cases. We would also like to point out that these cases start with high generative accuracy and the \emph{relative} differential in discrimination is quite minimal. 
All these observations lend support for \PHENOMENON. 

Instruction fine-tuned models went through both instruction-tuning and RLHF alignment while the base models are only pre-trained with the autoregressive objective.
It is reasonable to expect that instruction-tuned models would exhibit better performance in the discrimination phase as instruction-tuning is shown to make models better at solving a variety of tasks. Furthermore, classification tasks (that resemble our discrimination setup) are well-represented in most instruction-tuning datasets~\citep {wang2022benchmarking,Bach2022PromptSourceAI,longpre2023flan}. 
However, our empirical findings do not support it. 


\paragraph{Stronger models tend to be better at discrimination (larger \GAP{}).}
Our research has observed an interesting trend: an increase in \GAP{} seems to correlate with model capacity. This pattern is particularly pronounced among models in the same category (base models, fine-tuned models, and proprietary models developed by OpenAI). 
We also want to emphasize that some of the strongest models we tested—specifically LLaMa-3-70B, LLaMa-3-70B-Instruct, GPT-3.5-turbo, and GPT-4-turbo—show a positive \GAP{} across nearly all evaluated tasks, though the gap remains small enough for \PHENOMENON{} to still hold. We hypothesize that this is because weaker models have limited discrimination capabilities. Similar observations on the weaker models' discrimination capability have been reported in other studies \citep{saunders2022selfcritiquingmodelsassistinghuman, kadavath2022languagemodelsmostlyknow}.

\setlength\tabcolsep{0pt} 
\begin{table*}
    \centering
    \small
\scalebox{0.95}{
\begin{tabular}
{
@{\hspace{-1pt}}
  @{\hspace{-1pt}}
  >{\raggedright\arraybackslash}p{3.0cm} 
  @{\hspace{-5pt}} 
  >{\centering\arraybackslash}p{2.2cm}   
  >{\centering\arraybackslash}p{0.75cm}   
  >{\centering\arraybackslash}p{2.2cm}   
  >{\centering\arraybackslash}p{0.75cm}   
  >{\centering\arraybackslash}p{2.2cm}   
  >{\centering\arraybackslash}p{0.75cm}   
  >{\centering\arraybackslash}p{2.2cm}   
  >{\centering\arraybackslash}p{0.75cm}   
  @{}
}

\toprule
 & \multicolumn{2}{c}{\textbf{GSM8K}} & \multicolumn{2}{c}{\textbf{TriviaQA}} & \multicolumn{2}{c}{\textbf{MT-Bench}} & \multicolumn{2}{c}{\textbf{TruthfulQA}} \\
 \cmidrule(lr){2-3} \cmidrule(lr){4-5} \cmidrule(lr){6-7} \cmidrule(lr){8-9} 
 & \textbf{\GAP{}}  & \textbf{p-val} & \textbf{\GAP{}} & \textbf{p-val} & \textbf{\GAP{}} & \textbf{p-val} & \textbf{\GAP{}} & \textbf{p-val} \\ \cmidrule{2-9}
 \textbf{LLaMA-2 7B} & -0.6$_{(9.2\rightarrow 8.6)}$  & -- & -16.9$_{(37.1\rightarrow 20.2)}$  & -- & -0.09$_{(3.34\rightarrow 3.25)}$ & -- & -4.7$_{(30.5\rightarrow 25.8)}$ & -- \\
 \textbf{LLaMA-2 13B} & 0.0$_{(16.8\rightarrow 16.8)}$ & \fontsize{8.0pt}{10.5pt}\selectfont{0.50}  & 1.4$_{(45.2\rightarrow 46.6)}$ & \fontsize{8.0pt}{10.5pt}\selectfont{0.07} & -0.12$_{(4.15\rightarrow 4.03)}$ & -- & 2.1$_{(26.8\rightarrow 28.9)}$ & \fontsize{8.0pt}{10.5pt}\selectfont{0.10} \\
\textbf{LLaMA-2 70B} & 2.2$_{(44.0\rightarrow 46.2)}$ & \fontsize{8.0pt}{10.5pt}\selectfont{0.12} & 3.2$_{(53.2\rightarrow 56.4)}$ & \fontsize{8.0pt}{10.5pt}\selectfont{\textcolor{red}{0.00}}  & -0.12$_{(4.87\rightarrow 4.75)}$  & -- & 0.5$_{(28.9\rightarrow 29.4)}$ & \fontsize{8.0pt}{10.5pt}\selectfont{0.40} \\
 \textbf{LLaMA-3 8B} & -3.6$_{(38.6\rightarrow 35.0)}$  & -- & -2.3$_{(45.4\rightarrow 43.1)}$ & -- & 0.06$_{(5.47\rightarrow 5.53)}$ & \fontsize{8.0pt}{10.5pt}\selectfont{0.42} &  0.2$_{(27.2\rightarrow 27.4)}$ & \fontsize{8.0pt}{10.5pt}\selectfont{0.47} \\
 \textbf{LLaMA-3 70B} & 1.1$_{(77.7\rightarrow 78.8)}$  & \fontsize{8.0pt}{10.5pt}\selectfont{0.25} & 1.1$_{(64.2\rightarrow 65.3)}$ & \fontsize{8.0pt}{10.5pt}\selectfont{0.09}  & 0.14$_{(6.32\rightarrow 6.46)}$ & \fontsize{8.0pt}{10.5pt}\selectfont{0.36} & {0.8}$_{(36.8\rightarrow 37.6)}$ & \fontsize{8.0pt}{10.5pt}\selectfont{0.37} \\
\cmidrule{1-9}
 $\textbf{Phi-3-3.8B}_\textbf{Instruct}$ & -0.2$_{(77.9\rightarrow 77.7)}$ & -- & 0.7$_{(22.1\rightarrow 22.9)}$ & \fontsize{8.0pt}{10.5pt}\selectfont{0.11} &  -0.08$_{(7.33\rightarrow 7.25)}$ & -- & -0.2$_{(26.3\rightarrow 26.1)}$ & --\\
 $\textbf{LLaMA-2 7B}_\textbf{Chat}$ & -2.8$_{(20.4\rightarrow 17.6)}$ & -- &  -0.1$_{(16.1\rightarrow 16.0)}$  & -- & -0.13$_{(5.45\rightarrow 5.32)}$  & -- & 1.4$_{(48.8\rightarrow 50.2)}$ & \fontsize{8.0pt}{10.5pt}\selectfont{0.20} \\
 $\textbf{LLaMA-2 13B}_\textbf{Chat}$ & -5.5$_{(28.3\rightarrow 22.8)}$ & -- &  0.0$_{(25.5\rightarrow 25.5)}$  & -- & -0.51$_{(5.67\rightarrow 5.16)}$  & -- & -0.1$_{(44.9\rightarrow 44.8)}$ & -- \\
 $\textbf{LLaMA-2 70B}_\textbf{Chat}$ & -5.9$_{(42.5\rightarrow 36.6)}$ & -- &  -1.6$_{(47.8\rightarrow 46.2)}$  & -- & -0.17$_{(6.65\rightarrow 6.48)}$ & -- & 0.9$_{(48.6\rightarrow 49.5)}$ & \fontsize{8.0pt}{10.5pt}\selectfont{0.31} \\
 $\textbf{LLaMA-3 8B}_\textbf{Instruct}$ & 1.0$_{(76.9\rightarrow 77.9)}$ & \fontsize{8.0pt}{10.5pt}\selectfont{0.29}  & 0.6$_{(48.7\rightarrow 49.3)}$ & \fontsize{8.0pt}{10.5pt}\selectfont{0.26} & 0.14$_{(8.01\rightarrow 8.15)}$ & \fontsize{8.0pt}{10.5pt}\selectfont{0.32} &  -0.6$_{(50.1\rightarrow 49.5)}$ & --\\
 $\textbf{LLaMA-3 70B}_\textbf{Instruct}$ & 0.6$_{(92.2\rightarrow 92.8)}$ & \fontsize{8.0pt}{10.5pt}\selectfont{0.31}  & 1.1$_{(64.2\rightarrow 65.3)}$ & \fontsize{8.0pt}{10.5pt}\selectfont{0.11}  & 0.19$_{(8.60\rightarrow8.79)}$ & \fontsize{8.0pt}{10.5pt}\selectfont{0.23} & -1.1$_{(56.2\rightarrow 55.1)}$ & -- \\
 $\textbf{Mixtral-8x7B}_\textbf{Instruct}$ & 1.3$_{(59.6\rightarrow 60.9)}$ & \fontsize{8.0pt}{10.5pt}\selectfont{0.37}  & -3.4$_{(58.8\rightarrow 55.4)}$ & -- & -0.20$_{(8.39\rightarrow 8.19)}$ & -- & -0.4$_{(61.1\rightarrow 60.7)}$ & -- \\
 \textbf{GPT-3.5-turbo} & 1.1$_{(75.3\rightarrow 76.4)}$ & \fontsize{8.0pt}{10.5pt}\selectfont{0.37}  & {2.1$_{(67.1\rightarrow 69.2)}$} & \fontsize{8.0pt}{10.5pt}\selectfont{\textcolor{red}{0.01}}  & 0.17$_{(8.44\rightarrow 8.61)}$ & \fontsize{8.0pt}{10.5pt}\selectfont{0.26} & 0.4$_{(65.7\rightarrow 66.1)}$ & \fontsize{8.0pt}{10.5pt}\selectfont{0.41} \\
 \textbf{GPT-4-turbo} & 0.7$_{(93.6\rightarrow 94.3)}$ & \fontsize{8.0pt}{10.5pt}\selectfont{0.39}  & 0.2$_{(79.9\rightarrow 80.1)}$ & \fontsize{8.0pt}{10.5pt}\selectfont{0.40}  & --  & -- & 1.7$_{(77.4\rightarrow 79.1)}$ & \fontsize{8.0pt}{10.5pt}\selectfont{0.09} \\
\cmidrule{1-9}
 \textbf{Task Avg.} & -0.76 &  & -0.99  &  & -0.06 &  & 0.06 &  \\ 
\bottomrule
\end{tabular}
}
    \caption{
    Performance change defined as $\GAP{} := S_\text{disc} - S_\text{gen}$, with p-values indicating the likelihood that the observed difference is due to chance. 
    The generation performance and discriminative performance are shown as subscript: $(S_\text{gen}~\rightarrow~S_\text{disc})$.
    p-values are calculated only when \GAP{} $\geq 0$ because our one-sided hypothesis \PHENOMENON{} can only be rejected when $S_\text{disc}$ is equal to or greater than $S_\text{gen}$. A red p-value
    signifies a value less than 0.05.
    \textbf{For the majority of our results, $\GAP{}$ is small or negative, indicating similar or worse performance in the discrimination phase. The p-value for 54/56 experiments is less than 0.05, meaning \PHENOMENON{} is not rejected.}}
    \label{tab:main_result}
\end{table*}

\setlength\tabcolsep{6pt} 

\section{Further Analysis of \PHENOMENON} \label{sec:explain}
In this section, we outline experiments designed to provide further analysis of \PHENOMENON{}.

\subsection{Better Discrimination via Prompt-Engineering}
\label{sec:additional}
One might argue our current prompting setup doesn't fully capitalize on the model's capacity for discrimination. 
To make sure \PHENOMENON{} isn't an artifact of poor prompt engineering, we conduct additional experiments with LLaMA-2 Chat models on GSM8K, TriviaQA, and MT-Bench as their DG-DIFF on those tasks is mostly negative.

\paragraph{More in-context learning examples helps discrimination, though \GAP{} remains small or negative.}
Increasing the number of in-context learning (ICL) demonstrations is shown to improve performance~\citep{brown2020language}.
Is it possible that increasing the number of ICL examples in the discrimination phase will improve it, so much that \GAP{} becomes consistently positive?
To evaluate the effect of increasing ICL examples, we conduct experiments where the number of ICL examples (\#Shots) during the discrimination stage is doubled or tripled relative to the baseline in \autoref{tab:setup}. Note that we didn't triple the number of ICL examples for MT-Bench because it exceeds the context length for LLaMa-2 Chat models (4096 tokens).
The results, presented in \autoref{tab:ablation}, indicate that while increasing the number of ICL examples tends to increase \GAP{}, it remains small or negative. Furthermore, the performance improvement from adding ICL examples does not exhibit a consistent monotonic trend.

\paragraph{Chain-of-thought rational shows minimal impact on \GAP.} Recently,  \citet{stechly2024selfverification} pointed out that LLM evaluation also involves multi-step reasoning. To help with the reasoning in the discrimination phase, we add chain-of-thought rationals in the few-shot examples while keeping the number of examples constant. For GSM8K, 
we do not report anything since our default evaluation already contains rationales for answer selection. 
For TriviaQA, the CoT rationales explain the logic behind choosing an option. 
For MT-Bench, we supplement explanations for preferring one answer over another. A comparison between our prompts (w/ and w/o CoT rationales) is available in \autoref{app: COT}. As shown in  \autoref{tab:ablation}, the inclusion of CoT rational only shows minimal impact.

\begin{table}[h]
    \centering
    \small
    \setlength{\tabcolsep}{3pt}
    \begin{tabular}{lccc ccc ccc}
        \toprule
        \textbf{Model} & \multicolumn{3}{c}{\textbf{LLaMA-2 7B Chat}} & \multicolumn{3}{c}{\textbf{LLaMA-2 13B Chat}} & \multicolumn{3}{c}{\textbf{LLaMA-2 70B Chat}} \\
        \cmidrule(lr){2-4} \cmidrule(lr){5-7} \cmidrule(lr){8-10}
        \textbf{Setup}  & \textcolor{purple}{\textbf{+2$\times$\#ICL}} & \textcolor{purple}{\textbf{+3$\times$\#ICL}} & \textcolor{purple}{\textbf{+CoT}}  & \textcolor{purple}{\textbf{+2$\times$\#ICL}}& \textcolor{purple}{\textbf{+3$\times$\#ICL}} & \textcolor{purple}{\textbf{+CoT}} &  \textcolor{purple}{\textbf{+2$\times$\#ICL}} & \textcolor{purple}{\textbf{+3$\times$\#ICL}} & \textcolor{purple}{\textbf{+CoT}} \\
        \midrule
        \textbf{GSM8K}  & \textbf{-1.4} & 0.1 & -  & \textbf{-5.9} & \textbf{-6.8} & -  & \textbf{-5.8} & \textbf{-3.9} & - \\
        \textbf{TriviaQA}  & \textbf{-0.4} & 0.2 & \textbf{-0.3}  & 0.1 & 0.1 & \textbf{-0.3}  & \textbf{-1.7} & \textbf{-0.5} & \textbf{-1.8} \\
        \textbf{MT-Bench}  & \textbf{-0.09} & - & \textbf{-0.06}  & \textbf{-0.53} & - & \textbf{-0.41}  & \textbf{-0.19} & - & \textbf{-0.18} \\
        \bottomrule
    \end{tabular}
    \caption{
    \GAP{} upon various modifications with LLaMA-2 Chat models. 
    ``+~2~$\times$~\#ICL'' means doubling the number of in-context demonstrations during the discrimination phase. ``+~3~$\times$~\#ICL'' means tripling the number of in-context demonstrations. ``+~CoT'' stands for adding Chain-of-Thought rationales for the few-shot examples. \textbf{Extra prompt-engineering techniques during discrimination do not consistently close the performance gap.
    }}
    \label{tab:ablation}
\end{table}

\subsection{The Role of Objectives: Does Autoregressive Pre-training Explain our Results?}
\label{sec:pre-training objective}
The majority of modern LLMs are pre-trained with an autoregressive objective. Recent studies suggest that autoregressive objectives used during pre-training may have unexpected impacts on LLM behavior~\citep{ember}. Since the pre-training process of autoregressive models is more similar to generation than discrimination, we hypothesize \textit{\PHENOMENON{} is also partially caused by the use of autoregressive pre-training objective}.

To test this hypothesis, we evaluate Flan-T5-XXL (11B) and Flan-UL2 (20B) on the same tasks listed in \autoref{tab:main_result}, as these are the only prominent open-source non-autoregressive models available to the best of our knowledge. Flan-T5-XXL is pre-trained using a span corruption objective, where the loss is only calculated on the corrupted span \citep{raffel2020exploring}. Flan-UL2~\citep{flan} is pre-trained using mixture-of-denoisers that combines multiple denoising objective functions.
Our findings, detailed in \autoref{tab:flan}, reveal their \GAP{} across all tasks are positive except for Flan-T5-XXL on MT-Bench. In fact, both models exhibit significantly higher \GAP{} and even more significantly higher \emph{relative} \GAP{} compared to the autoregressive models we tested in \autoref{tab:main_result}. Moreover, for both TriviaQA and TruthfulQA, the \PHENOMENON{} hypothesis is rejected. This outcome lends empirical support to the hypothesis that \PHENOMENON{} could be related to autoregressive pre-training.

\begin{table*}[h]
    \centering
    \small
\begin{tabular}{@{}l|cccc}
\toprule
& \multicolumn{4}{c}{$\textbf{\GAP{}} _{(\textbf{S}_\textbf{gen}~\rightarrow~\textbf{S}_\textbf{disc})}$} \\ 
\cmidrule{2-5}
 & \textbf{GSM8K} & \textbf{TriviaQA} & \textbf{MT-Bench} & \textbf{TruthfulQA} \\
\textbf{Flan-T5 XXL} & {1.3}$_{(13.3\rightarrow 14.4)}$ & 5.8$_{(28.7\rightarrow 34.5)}$ & \textbf{-0.06}$_{(2.02\rightarrow 1.96)}$ & {6.0}$_{(20.1\rightarrow 26.1)}$ \\
\textbf{Flan-UL2} & 0.5$_{(21.6\rightarrow 22.1)}$ & 4.2$_{(52.7\rightarrow 56.9)}$ & {0.16}$_{(1.98\rightarrow 2.14)}$ & 4.8$_{(31.3\rightarrow 36.1)}$\\
\midrule
\textbf{Task Avg.} & 0.9 & 5.0 & 0.05 & 5.4 \\
\bottomrule
\end{tabular}
    \caption{Flan-T5-XXL and Flan-UL2 tested on the same setup as  \autoref{tab:main_result}. \GAP{} for all models across all tasks are positive except for Flan-T5-XXL on MT-Bench. \textbf{Both models demonstrate significantly higher average \GAP{} compared to autoregressive models.}}
    \label{tab:flan}
\end{table*}

It is also important to note that the pre-training processes for these two model classes differ from autoregressively
pre-trained counterparts beyond the objective function. For example, Flan-T5 is pre-trained with most inputs provided, except for the corrupted spans. Furthermore, the datasets used for pre-training these LLMs can vary. Therefore, when uncovering the underlying reason why \PHENOMENON{} does not occur on Flan-T5 and Flan-UL2, caution should be exercised before drawing definitive conclusions.

\subsection{Do Prior Findings in Self-Refinement Contradict \PHENOMENON{}?} \label{sec:self-reflection}
The process of self-refine involves utilizing the same LLM to provide feedback for its own generation and using the feedback to refine the generation. Both \citet{cant} and \citet{self-refine} suggested LLMs can self-refine on tasks other than reasoning. Does this contradict our assertions? 

We replicated the experiment outlined in \citet{self-refine} and observed the following: 

(1)  \textbf{For some evaluated tasks, certain aspects can be exploited for artificially amplifying task performance without actually improving with the feedback.
} 
For example, on the task of constrained generation, where the objective is to generate sentences containing specific keywords, self-refine with LLMs often leads to progressively longer sentences that simply extend previous ones. 
Thus, even if the refined sentences do not incorporate new keywords and continue to grow longer (often, ignoring the feedback from the prior round), the task performance still shows a monotonic improvement. A more detailed explanation of this behavior across additional tasks is provided in \autoref{tab:self-refine-problem}. To further illustrate our point, an example question and model output from the acronym generation task can be found in \autoref{example_acronym_generation} in \autoref{app:extra_analysis_on_self-refine}, and another example from the constraint generation task is presented in \autoref{example_constraint_generation} in the same appendix.

\begin{table*}[ht]
    \centering
    \small
    \setlength{\tabcolsep}{3pt}
    \begin{tabular}{lllc}
        \toprule
        \textbf{Task}& \textbf{Issues} & \textbf{Detailed Explanation}  & \textbf{Pref.\%} \\
        \midrule
        \multirow{2}{*}{Sentiment Reversal} & \multirow{2}{*}{Lack of Reasoning} &  Refinements simply make the   & \multirow{2}{*}{58.7\%}\\
        & & sentiment more and more positive & \\
        Dialogue Response  & \multirow{2}{*}{Reward Inconsistency} & Reward assigned by LLMs doesn't & \multirow{2}{*}{52.4\%}   \\
        Generation& & increase monotonically & \\
        Code Readability & \multirow{2}{*}{Lack of Reasoning}  &  Refinements simply makes variable   & \multirow{2}{*}{53.3\%}  \\
        Improvement& & names longer and more descriptive & \\
        \multirow{2}{*}{Acronym Generation} & \multirow{2}{*}{Reward Inconsistency} & Reward assigned by LLMs doesn't   & \multirow{2}{*}{46.5\%}\\
        & & increase monotonically & \\
       \multirow{2}{*}{Constrained Generation} & \multirow{2}{*}{Lack of Reasoning} & Refinements simply extends &\multirow{2}{*}{54.7\%}\\
       & & previous generation & \\
        \bottomrule
    \end{tabular}
    \caption{
    {\textbf{Explanation for some of the issues on tasks that \citet{self-refine} tested and the percentage of times the model prefers self-refined subsequent generations than previous generations.} GSM8K isn't included here because it didn't get much improvement through self-refine in the original paper. Code optimization isn't included either due to the complexity of running experiments.}
    }
    \label{tab:self-refine-problem}
\end{table*}

(2) \textbf{For some evaluated tasks, the evaluation score assigned by the model for each iteration of self-refine is not monotonically increasing.} We use the same model involved in self-refinement to evaluate each refined output. Ideally, the scores would improve with refinement, but for tasks like acronym generation and dialogue response, the model often assigns lower scores to refined outputs. This suggests that the observed improvement may be due to lower initial output quality, as noted in \citet{cant}.

(3) 
\textbf{Quantifying the percentage of times models prefer self-refined subsequent generations to the previous generation, a marginal preference for self-refined generation was observed.}
We used the same models to discriminate between previous generations and self-refined subsequent generations for tasks referenced in \citet{self-refine}, thus extending the evaluation of \PHENOMENON{} to a broader range of real-world tasks beyond reasoning.
Our results in \autoref{tab:self-refine-problem} indicate that models prefer self-refined generations only around 54\% of the time, meaning on those tasks LLMs are still not consistently better at discriminating among previously-generated alternatives than generating initial responses.

\section{Further Discussion} \label{sec:implication}

\paragraph{\PHENOMENON{} likely poses a barrier for continued progress in self-rewarding LLMs.}
\label{sec:rlaif}
Few recent works like Self-Rewarding Language Models \citep{self-rewarding} generate preference pairs consisting of an instruction prompt $x$, a winning response $y\_win$, and a losing response $y\_lose$ to facilitate self-reward for instruction-following fine-tuning. While their setup also involves discriminating between previously generated outputs, our findings do not challenge its effectiveness. Their method selects winning and losing pairs from a larger set of generations, with the key factor being the correct ordering of these pairs. This setup simplifies the task, especially when straightforward heuristics like choosing the highest and lowest-scoring responses exist. In contrast, our approach requires the model to identify the single best generation, demanding a much finer level of granularity.

However, an interesting pattern from~\citet{self-rewarding} is that there seem to be diminishing returns after a few iterations of self-rewarding. We hypothesize this may be linked to \PHENOMENON{} because if the overall ability of LLMs to discriminate is inferior to their ability to generate, it becomes challenging to engage in a virtuous cycle that simultaneously enhances the model's capability to follow instructions and generate self-rewards.

\paragraph{Controlled Modification of Experimental Setting with Simplified Distractors} 
\label{sec:simple_setup}
Here we consider the extent to which \PHENOMENON{} may hold.
For example, is \PHENOMENON{} potentially a fundamental limitation of LLMs pre-trained with an autoregressive objective, 
or can a change in data distribution alter the outcome? 
To address this question, we conduct experiments in an unconventional setting that simplifies the discrimination phase by substituting incorrect candidates with simpler ones for discrimination. 

The experiments are conducted on TriviaQA and GSM8K. 
TriviaQA contains a wide range of answer categories, including names, locations, historical events, etc. 
For this dataset, we simplify the discrimination phase by substituting incorrect answer generations for question $A$ with correct answer generations from another question $B$ (left panel in  \autoref{fig:unlikely_setup_fig},  \autoref{app:explanation_for_simplified_discrimination_phase}). 
As for GSM8K, we create 
simplified distractors by randomly multiplying or dividing incorrect generated answers by $100$ (right panel in  \autoref{fig:unlikely_setup_fig},  \autoref{app:explanation_for_simplified_discrimination_phase}).

\autoref{fig:rand} in  \autoref{app:explanation_for_simplified_discrimination_phase} clearly shows that simplifying the incorrect candidates improves \GAP. 
For TriviaQA, ${S}_{\text{disc}}$ exceeds ${S}_{\text{gen}}$ by a large margin.
For GSM8K, all models tested also demonstrate improved \GAP{}.

\section{Limitations}
\paragraph{Challenges in controlled study of LLMs in relation to \PHENOMENON.}
One limitation of our research stems from the
difficulty in measuring
the impact of pre-training data and pre-training objectives. The vast amount of pre-training data makes it hard to evaluate its effect, leaving 
important aspects underexplored.

\paragraph{Potential influence of lengthier discrimination prompt on \PHENOMENON{}.} The prompt used in the discrimination phase is inherently lengthier than the generation prompt as it also includes the generated candidate answers. This increase in length may pose challenges to the model's processing capabilities. Investigating the impact of prompt length is complex as simply adding superfluous content to lengthen the generation or discrimination prompt might unintentionally influence the outcomes. Therefore, we highlight this area for further exploration to better understand the implications of prompt length for \PHENOMENON{}.

\paragraph{Limitations in experimental scope}
Another limitation of our study is the scope of our experiments. While we tested \PHENOMENON{} across multiple tasks and domains using prominent LLMs, expanding to more models and tasks could further validate \PHENOMENON{}.

We want to note while the current results support \PHENOMENON{}, we are \emph{not} claiming that LLMs can \emph{never} be better at discrimination than generation. Some studies \citep{welleck2022selfcorrect, chen2024gainingwisdomsetbacksaligning} suggest that fine-tuning specifically on refinement data can improve discrimination capabilities, though this may come at the expense of the model's generality. Whether it is possible to train a model that maintains generality while excelling at discrimination remains an open research question. 

In addition, our discrimination setup is designed to be simple, allowing our method to be directly applied to a wide range of tasks while helping us better understand the inherent characteristics and challenges of LLMs. Exploring more complex techniques like problem decomposition and answer verification to enhance discrimination is beyond the scope of this paper.

\paragraph{Challenges in determining model's preference} Determining a model's preference over several candidate generations can be challenging due to various biases \citep{alzahrani2024benchmarkstargetsrevealingsensitivity, wang2024mmluprorobustchallengingmultitask}. Following the methodology used in other self-improvement studies~\citep{self-rewarding}, we employ \textit{LLM-as-a-Judge prompting}~\citep{zheng2023judging} to elicit answer choice from the model. It is conceivable that the LLMs we examine can be biased toward certain answer options or different answer formats (e.g., labels of A/B/C/D or [1]/[2]/[3]/[4]). Another method that we did not explore is ranking candidate answers based on the LLM's assigned probability for each answer text. However, it is worth noting that this approach can also be biased by factors like the text fluency from the pre-training data.

\paragraph{Limited focus on other stages of self-improvement} Self-improvement involves multiple stages, such as self-discrimination, critique generation, and generating additional answers after self-evaluation. Our focus is mainly on the first stage, with less emphasis on others. However, it is generally agreed \citep{cant, can_correct} that for LLMs to succeed at other stages reliably, they must first excel at the initial self-discrimination stage. 

Even within the first stage, approaches can vary; some generate a single answer and decide if another is needed, whereas our work explores generating multiple answers followed by discrimination. However, we believe that overall success in this stage is ultimately dependent on the model's discrimination capability.

\section{Conclusion}
We focused on the question of whether language models are strictly better at discriminating their prior generations vs. generating responses directly. 
We proposed a metric for comparing these capabilities and used it to evaluate several current LLMs.  For those models and tasks, we do not observe that discrimination is reliably better than generation, in fact, we often observed it was worse. These results raise concerns about the potential for LLM self-improvement on \textit{\textbf{any}} task.

\bibliography{colm2024_conference}

\providecommand{\CNFX}[1]{{\em{\textrm{(#1)}}}}
\begin{thebibliography}{55}
\providecommand{\natexlab}[1]{#1}
\providecommand{\url}[1]{\texttt{#1}}
\expandafter\ifx\csname urlstyle\endcsname\relax
  \providecommand{\doi}[1]{doi: #1}\else
  \providecommand{\doi}{doi: \begingroup \urlstyle{rm}\Url}\fi

\bibitem[Abdin et~al.(2024)Abdin, Jacobs, Awan, Aneja, Awadallah, Awadalla, Bach, Bahree, Bakhtiari, Bao, Behl, Benhaim, Bilenko, Bjorck, Bubeck, Cai, Cai, Mendes, Chen, Chaudhary, Chen, Chen, Chen, Chen, Chopra, Dai, Giorno, de~Rosa, Dixon, Eldan, Fragoso, Iter, Gao, Gao, Gao, Garg, Goswami, Gunasekar, Haider, Hao, Hewett, Huynh, Javaheripi, Jin, Kauffmann, Karampatziakis, Kim, Khademi, Kurilenko, Lee, Lee, Li, Li, Liang, Liden, Liu, Liu, Liu, Lin, Lin, Luo, Madan, Mazzola, Mitra, Modi, Nguyen, Norick, Patra, Perez-Becker, Portet, Pryzant, Qin, Radmilac, Rosset, Roy, Ruwase, Saarikivi, Saied, Salim, Santacroce, Shah, Shang, Sharma, Shukla, Song, Tanaka, Tupini, Wang, Wang, Wang, Wang, Ward, Wang, Witte, Wu, Wyatt, Xiao, Xu, Xu, Xu, Yadav, Yang, Yang, Yang, Yang, Yu, Yuan, Zhang, Zhang, Zhang, Zhang, Zhang, Zhang, Zhang, and Zhou]{abdin2024phi3technicalreporthighly}
Marah Abdin, Sam~Ade Jacobs, Ammar~Ahmad Awan, Jyoti Aneja, Ahmed Awadallah, Hany Awadalla, Nguyen Bach, Amit Bahree, Arash Bakhtiari, Jianmin Bao, Harkirat Behl, Alon Benhaim, Misha Bilenko, Johan Bjorck, Sébastien Bubeck, Qin Cai, Martin Cai, Caio César~Teodoro Mendes, Weizhu Chen, Vishrav Chaudhary, Dong Chen, Dongdong Chen, Yen-Chun Chen, Yi-Ling Chen, Parul Chopra, Xiyang Dai, Allie~Del Giorno, Gustavo de~Rosa, Matthew Dixon, Ronen Eldan, Victor Fragoso, Dan Iter, Mei Gao, Min Gao, Jianfeng Gao, Amit Garg, Abhishek Goswami, Suriya Gunasekar, Emman Haider, Junheng Hao, Russell~J. Hewett, Jamie Huynh, Mojan Javaheripi, Xin Jin, Piero Kauffmann, Nikos Karampatziakis, Dongwoo Kim, Mahoud Khademi, Lev Kurilenko, James~R. Lee, Yin~Tat Lee, Yuanzhi Li, Yunsheng Li, Chen Liang, Lars Liden, Ce~Liu, Mengchen Liu, Weishung Liu, Eric Lin, Zeqi Lin, Chong Luo, Piyush Madan, Matt Mazzola, Arindam Mitra, Hardik Modi, Anh Nguyen, Brandon Norick, Barun Patra, Daniel Perez-Becker, Thomas Portet, Reid Pryzant, Heyang
  Qin, Marko Radmilac, Corby Rosset, Sambudha Roy, Olatunji Ruwase, Olli Saarikivi, Amin Saied, Adil Salim, Michael Santacroce, Shital Shah, Ning Shang, Hiteshi Sharma, Swadheen Shukla, Xia Song, Masahiro Tanaka, Andrea Tupini, Xin Wang, Lijuan Wang, Chunyu Wang, Yu~Wang, Rachel Ward, Guanhua Wang, Philipp Witte, Haiping Wu, Michael Wyatt, Bin Xiao, Can Xu, Jiahang Xu, Weijian Xu, Sonali Yadav, Fan Yang, Jianwei Yang, Ziyi Yang, Yifan Yang, Donghan Yu, Lu~Yuan, Chengruidong Zhang, Cyril Zhang, Jianwen Zhang, Li~Lyna Zhang, Yi~Zhang, Yue Zhang, Yunan Zhang, and Xiren Zhou.
\newblock Phi-3 technical report: A highly capable language model locally on your phone, 2024.
\newblock URL \url{https://arxiv.org/abs/2404.14219}.

\bibitem[Alexander(2003)]{expertise1}
Patricia~A Alexander.
\newblock The development of expertise: The journey from acclimation to proficiency.
\newblock In \emph{Educational researcher}, 2003.
\newblock URL \url{https://www.jstor.org/stable/3700080}.

\bibitem[Alzahrani et~al.(2024)Alzahrani, Alyahya, Alnumay, Alrashed, Alsubaie, Almushaykeh, Mirza, Alotaibi, Altwairesh, Alowisheq, Bari, and Khan]{alzahrani2024benchmarkstargetsrevealingsensitivity}
Norah Alzahrani, Hisham~Abdullah Alyahya, Yazeed Alnumay, Sultan Alrashed, Shaykhah Alsubaie, Yusef Almushaykeh, Faisal Mirza, Nouf Alotaibi, Nora Altwairesh, Areeb Alowisheq, M~Saiful Bari, and Haidar Khan.
\newblock When benchmarks are targets: Revealing the sensitivity of large language model leaderboards, 2024.
\newblock URL \url{https://arxiv.org/abs/2402.01781}.

\bibitem[Arora \& Kambhampati(2023)Arora and Kambhampati]{arora2023learning}
Daman Arora and Subbarao Kambhampati.
\newblock Learning and leveraging verifiers to improve planning capabilities of pre-trained language models.
\newblock \emph{CoRR}, 2023.
\newblock URL \url{https://arxiv.org/abs/2305.17077}.

\bibitem[Bach et~al.(2022)Bach, Sanh, Yong, Webson, Raffel, Nayak, Sharma, Kim, Bari, F{\'e}vry, Alyafeai, Dey, Santilli, Sun, Ben-David, Xu, Chhablani, Wang, Fries, Al-shaibani, Sharma, Thakker, Almubarak, Tang, Jiang, and Rush]{Bach2022PromptSourceAI}
Stephen~H. Bach, Victor Sanh, Zheng~Xin Yong, Albert Webson, Colin Raffel, Nihal~V. Nayak, Abheesht Sharma, Taewoon Kim, M~Saiful Bari, Thibault F{\'e}vry, Zaid Alyafeai, Manan Dey, Andrea Santilli, Zhiqing Sun, Srulik Ben-David, Canwen Xu, Gunjan Chhablani, Han Wang, Jason~Alan Fries, Maged~S. Al-shaibani, Shanya Sharma, Urmish Thakker, Khalid Almubarak, Xiangru Tang, Mike Tian-Jian Jiang, and Alexander~M. Rush.
\newblock Promptsource: An integrated development environment and repository for natural language prompts.
\newblock \emph{ArXiv}, abs/2202.01279, 2022.
\newblock URL \url{https://arxiv.org/abs/2202.01279}.

\bibitem[Brown et~al.(2020)Brown, Mann, Ryder, Subbiah, Kaplan, Dhariwal, Neelakantan, Shyam, Sastry, Askell, et~al.]{brown2020language}
Tom Brown, Benjamin Mann, Nick Ryder, Melanie Subbiah, Jared~D Kaplan, Prafulla Dhariwal, Arvind Neelakantan, Pranav Shyam, Girish Sastry, Amanda Askell, et~al.
\newblock Language models are few-shot learners.
\newblock \emph{Advances in Neural Information Processing Systems \CNFX{NeurIPS}}, 2020.
\newblock URL \url{https://arxiv.org/abs/2005.14165}.

\bibitem[Butt et~al.(2024)Butt, Manczak, Wiggers, Rainone, Zhang, Defferrard, and Cohen]{butt2024codeit}
Natasha Butt, Blazej Manczak, Auke Wiggers, Corrado Rainone, David Zhang, Michaël Defferrard, and Taco Cohen.
\newblock Codeit: Self-improving language models with prioritized hindsight replay.
\newblock \emph{arXiv preprint arXiv:2402.04858}, 2024.
\newblock URL \url{https://arxiv.org/abs/2402.04858}.

\bibitem[Chen et~al.(2024{\natexlab{a}})Chen, Wang, Yang, Han, Hong, Mi, Xu, Liu, Huang, Li, Yeung, Shang, Jiang, and Liu]{chen2024gainingwisdomsetbacksaligning}
Kai Chen, Chunwei Wang, Kuo Yang, Jianhua Han, Lanqing Hong, Fei Mi, Hang Xu, Zhengying Liu, Wenyong Huang, Zhenguo Li, Dit-Yan Yeung, Lifeng Shang, Xin Jiang, and Qun Liu.
\newblock Gaining wisdom from setbacks: Aligning large language models via mistake analysis, 2024{\natexlab{a}}.
\newblock URL \url{https://arxiv.org/abs/2310.10477}.

\bibitem[Chen et~al.(2024{\natexlab{b}})Chen, White, Mooney, Payani, Su, and Sun]{chen2024treesearchusefulllm}
Ziru Chen, Michael White, Raymond Mooney, Ali Payani, Yu~Su, and Huan Sun.
\newblock When is tree search useful for llm planning? it depends on the discriminator, 2024{\natexlab{b}}.
\newblock URL \url{https://arxiv.org/abs/2402.10890}.

\bibitem[Chen et~al.(2024{\natexlab{c}})Chen, Deng, Yuan, Ji, and Gu]{self-play}
Zixiang Chen, Yihe Deng, Huizhuo Yuan, Kaixuan Ji, and Quanquan Gu.
\newblock Self-play fine-tuning converts weak language models to strong language models.
\newblock \emph{CoRR}, 2024{\natexlab{c}}.
\newblock URL \url{https://arxiv.org/abs/2401.01335}.

\bibitem[Chiang \& Lee(2023)Chiang and Lee]{DBLP:conf/acl/ChiangL23}
David~Cheng{-}Han Chiang and Hung{-}yi Lee.
\newblock Can large language models be an alternative to human evaluations?
\newblock In Anna Rogers, Jordan~L. Boyd{-}Graber, and Naoaki Okazaki (eds.), \emph{ACL}, 2023.
\newblock URL \url{https://arxiv.org/abs/2305.01937}.

\bibitem[Chung et~al.(2022)Chung, Hou, Longpre, Zoph, Tay, Fedus, Li, Wang, Dehghani, Brahma, Webson, Gu, Dai, Suzgun, Chen, Chowdhery, Narang, Mishra, Yu, Zhao, Huang, Dai, Yu, Petrov, Chi, Dean, Devlin, Roberts, Zhou, Le, and Wei]{flan}
Hyung~Won Chung, Le~Hou, Shayne Longpre, Barret Zoph, Yi~Tay, William Fedus, Eric Li, Xuezhi Wang, Mostafa Dehghani, Siddhartha Brahma, Albert Webson, Shixiang~Shane Gu, Zhuyun Dai, Mirac Suzgun, Xinyun Chen, Aakanksha Chowdhery, Sharan Narang, Gaurav Mishra, Adams Yu, Vincent~Y. Zhao, Yanping Huang, Andrew~M. Dai, Hongkun Yu, Slav Petrov, Ed~H. Chi, Jeff Dean, Jacob Devlin, Adam Roberts, Denny Zhou, Quoc~V. Le, and Jason Wei.
\newblock Scaling instruction-finetuned language models.
\newblock \emph{CoRR}, 2022.
\newblock URL \url{https://arxiv.org/abs/2210.11416}.

\bibitem[Cobbe et~al.(2021)Cobbe, Kosaraju, Bavarian, Chen, Jun, Kaiser, Plappert, Tworek, Hilton, Nakano, Hesse, and Schulman]{gsm8k}
Karl Cobbe, Vineet Kosaraju, Mohammad Bavarian, Mark Chen, Heewoo Jun, Lukasz Kaiser, Matthias Plappert, Jerry Tworek, Jacob Hilton, Reiichiro Nakano, Christopher Hesse, and John Schulman.
\newblock Training verifiers to solve math word problems, 2021.
\newblock URL \url{https://arxiv.org/abs/2110.14168}.

\bibitem[Corder(1967)]{corder1967significance}
Stephen~Pit Corder.
\newblock The significance of learner's errors.
\newblock 1967.
\newblock URL \url{https://eric.ed.gov/?id=ED019903}.

\bibitem[Dror et~al.(2018)Dror, Baumer, Shlomov, and Reichart]{dror2018hitchhiker}
Rotem Dror, Gili Baumer, Segev Shlomov, and Roi Reichart.
\newblock The hitchhiker’s guide to testing statistical significance in natural language processing.
\newblock In \emph{Proceedings of the 56th annual meeting of the association for computational linguistics (volume 1: Long papers)}, pp.\  1383--1392, 2018.
\newblock URL \url{https://aclanthology.org/P18-1128/}.

\bibitem[Gilardi et~al.(2023)Gilardi, Alizadeh, and Kubli]{Gilardi_2023}
Fabrizio Gilardi, Meysam Alizadeh, and Maël Kubli.
\newblock Chatgpt outperforms crowd workers for text-annotation tasks.
\newblock \emph{Proceedings of the National Academy of Sciences}, July 2023.
\newblock ISSN 1091-6490.
\newblock URL \url{https://arxiv.org/abs/2303.15056}.

\bibitem[Gou et~al.(2024)Gou, Shao, Gong, Shen, Yang, Duan, and Chen]{gou2024critic}
Zhibin Gou, Zhihong Shao, Yeyun Gong, Yelong Shen, Yujiu Yang, Nan Duan, and Weizhu Chen.
\newblock Critic: Large language models can self-correct with tool-interactive critiquing, 2024.
\newblock URL \url{https://arxiv.org/abs/2305.11738}.

\bibitem[He et~al.(2023)He, Zhang, Wang, Kumar, Cho, Glass, and Tsvetkov]{he-etal-2023-blind}
Tianxing He, Jingyu Zhang, Tianle Wang, Sachin Kumar, Kyunghyun Cho, James Glass, and Yulia Tsvetkov.
\newblock On the blind spots of model-based evaluation metrics for text generation.
\newblock In Anna Rogers, Jordan Boyd-Graber, and Naoaki Okazaki (eds.), \emph{ACL}, July 2023.
\newblock URL \url{https://arxiv.org/abs/2212.10020}.

\bibitem[Huang et~al.(2023)Huang, Chen, Mishra, Zheng, Yu, Song, and Zhou]{cant}
Jie Huang, Xinyun Chen, Swaroop Mishra, Huaixiu~Steven Zheng, Adams~Wei Yu, Xinying Song, and Denny Zhou.
\newblock Large language models cannot self-correct reasoning yet, 2023.
\newblock URL \url{https://arxiv.org/abs/2310.01798}.

\bibitem[Joshi et~al.(2017)Joshi, Choi, Weld, and Zettlemoyer]{TriviaQA}
Mandar Joshi, Eunsol Choi, Daniel~S. Weld, and Luke Zettlemoyer.
\newblock Triviaqa: {A} large scale distantly supervised challenge dataset for reading comprehension.
\newblock In \emph{ACL}, 2017.
\newblock URL \url{https://arxiv.org/abs/1705.03551}.

\bibitem[Kadavath et~al.(2022)Kadavath, Conerly, Askell, Henighan, Drain, Perez, Schiefer, Hatfield-Dodds, DasSarma, Tran-Johnson, Johnston, El-Showk, Jones, Elhage, Hume, Chen, Bai, Bowman, Fort, Ganguli, Hernandez, Jacobson, Kernion, Kravec, Lovitt, Ndousse, Olsson, Ringer, Amodei, Brown, Clark, Joseph, Mann, McCandlish, Olah, and Kaplan]{kadavath2022languagemodelsmostlyknow}
Saurav Kadavath, Tom Conerly, Amanda Askell, Tom Henighan, Dawn Drain, Ethan Perez, Nicholas Schiefer, Zac Hatfield-Dodds, Nova DasSarma, Eli Tran-Johnson, Scott Johnston, Sheer El-Showk, Andy Jones, Nelson Elhage, Tristan Hume, Anna Chen, Yuntao Bai, Sam Bowman, Stanislav Fort, Deep Ganguli, Danny Hernandez, Josh Jacobson, Jackson Kernion, Shauna Kravec, Liane Lovitt, Kamal Ndousse, Catherine Olsson, Sam Ringer, Dario Amodei, Tom Brown, Jack Clark, Nicholas Joseph, Ben Mann, Sam McCandlish, Chris Olah, and Jared Kaplan.
\newblock Language models (mostly) know what they know, 2022.
\newblock URL \url{https://arxiv.org/abs/2207.05221}.

\bibitem[Krishna(2023)]{krishna2023intersectionselfcorrectiontrustlanguage}
Satyapriya Krishna.
\newblock On the intersection of self-correction and trust in language models, 2023.
\newblock URL \url{https://arxiv.org/abs/2311.02801}.

\bibitem[Lin et~al.(2022)Lin, Hilton, and Evans]{lin-etal-2022-truthfulqa}
Stephanie Lin, Jacob Hilton, and Owain Evans.
\newblock {T}ruthful{QA}: Measuring how models mimic human falsehoods.
\newblock In Smaranda Muresan, Preslav Nakov, and Aline Villavicencio (eds.), \emph{ACL}. Association for Computational Linguistics, May 2022.
\newblock URL \url{https://arxiv.org/abs/2109.07958}.

\bibitem[Lin et~al.(2024)Lin, Gou, Liang, Luo, Liu, and Yang]{lin2024criticbenchbenchmarkingllmscritiquecorrect}
Zicheng Lin, Zhibin Gou, Tian Liang, Ruilin Luo, Haowei Liu, and Yujiu Yang.
\newblock Criticbench: Benchmarking llms for critique-correct reasoning, 2024.
\newblock URL \url{https://arxiv.org/abs/2402.14809}.

\bibitem[Liu et~al.(2023{\natexlab{a}})Liu, Pasunuru, Hajishirzi, Choi, and Celikyilmaz]{liu2023crystal}
Jiacheng Liu, Ramakanth Pasunuru, Hannaneh Hajishirzi, Yejin Choi, and Asli Celikyilmaz.
\newblock Crystal: Introspective reasoners reinforced with self-feedback.
\newblock \emph{arXiv preprint arXiv:2301.04921}, 2023{\natexlab{a}}.
\newblock URL \url{https://arxiv.org/abs/2310.04921}.

\bibitem[Liu et~al.(2023{\natexlab{b}})Liu, Fabbri, Chen, Zhao, Han, Joty, Liu, Radev, Wu, and Cohan]{DBLP:journals/corr/abs-2311-09184}
Yixin Liu, Alexander~R. Fabbri, Jiawen Chen, Yilun Zhao, Simeng Han, Shafiq Joty, Pengfei Liu, Dragomir Radev, Chien{-}Sheng Wu, and Arman Cohan.
\newblock Benchmarking generation and evaluation capabilities of large language models for instruction controllable summarization.
\newblock \emph{CoRR}, 2023{\natexlab{b}}.
\newblock URL \url{https://arxiv.org/abs/2311.09184}.

\bibitem[Longpre et~al.(2023)Longpre, Hou, Vu, Webson, Chung, Tay, Zhou, Le, Zoph, Wei, et~al.]{longpre2023flan}
Shayne Longpre, Le~Hou, Tu~Vu, Albert Webson, Hyung~Won Chung, Yi~Tay, Denny Zhou, Quoc~V Le, Barret Zoph, Jason Wei, et~al.
\newblock The flan collection: Designing data and methods for effective instruction tuning.
\newblock \emph{arXiv preprint arXiv:2301.13688}, 2023.
\newblock URL \url{https://arxiv.org/abs/2301.13688}.

\bibitem[Madaan et~al.(2023)Madaan, Tandon, Gupta, Hallinan, Gao, Wiegreffe, Alon, Dziri, Prabhumoye, Yang, Welleck, Majumder, Gupta, Yazdanbakhsh, and Clark]{self-refine}
Aman Madaan, Niket Tandon, Prakhar Gupta, Skyler Hallinan, Luyu Gao, Sarah Wiegreffe, Uri Alon, Nouha Dziri, Shrimai Prabhumoye, Yiming Yang, Sean Welleck, Bodhisattwa~Prasad Majumder, Shashank Gupta, Amir Yazdanbakhsh, and Peter Clark.
\newblock Self-refine: Iterative refinement with self-feedback.
\newblock \emph{CoRR}, 2023.
\newblock URL \url{https://arxiv.org/abs/2303.17651}.

\bibitem[Mayo(1996)]{mayo1996error}
Deborah~G Mayo.
\newblock \emph{Error and the growth of experimental knowledge}.
\newblock University of Chicago Press, 1996.
\newblock URL \url{https://errorstatistics.com/wp-content/uploads/2020/10/egek-pdf-red.pdf}.

\bibitem[McCoy et~al.(2023)McCoy, Yao, Friedman, Hardy, and Griffiths]{ember}
R.~Thomas McCoy, Shunyu Yao, Dan Friedman, Matthew Hardy, and Thomas~L. Griffiths.
\newblock Embers of autoregression: Understanding large language models through the problem they are trained to solve.
\newblock \emph{CoRR}, 2023.
\newblock URL \url{https://arxiv.org/abs/2309.13638}.

\bibitem[Mcnemar(1947)]{Mcnemar1947NoteOT}
Quinn Mcnemar.
\newblock Note on the sampling error of the difference between correlated proportions or percentages.
\newblock \emph{Psychometrika}, 12:\penalty0 153--157, 1947.
\newblock URL \url{https://api.semanticscholar.org/CorpusID:46226024}.

\bibitem[Pan et~al.(2023)Pan, Saxon, Xu, Nathani, Wang, and Wang]{Pan2023AutomaticallyCL}
Liangming Pan, Michael~Stephen Saxon, Wenda Xu, Deepak Nathani, Xinyi Wang, and William~Yang Wang.
\newblock Automatically correcting large language models: Surveying the landscape of diverse self-correction strategies.
\newblock \emph{ArXiv}, abs/2308.03188, 2023.
\newblock URL \url{https://api.semanticscholar.org/CorpusID:260682695}.

\bibitem[Raffel et~al.(2020)Raffel, Shazeer, Roberts, Lee, Narang, Matena, Zhou, Li, and Liu]{raffel2020exploring}
Colin Raffel, Noam Shazeer, Adam Roberts, Katherine Lee, Sharan Narang, Michael Matena, Yanqi Zhou, Wei Li, and Peter~J Liu.
\newblock Exploring the limits of transfer learning with a unified text-to-text transformer.
\newblock \emph{Journal of Machine Learning Research \CNFX{JMLR}}, 2020.
\newblock URL \url{https://arxiv.org/abs/1910.10683}.

\bibitem[Sadeqi~Azer et~al.(2020)Sadeqi~Azer, Khashabi, Sabhwawal, and Roth]{sadeqi-azer2020not}
Erfan Sadeqi~Azer, Daniel Khashabi, Ashish Sabhwawal, and Dan Roth.
\newblock Not all claims are created equal: Choosing the right approach to assess your hypotheses.
\newblock In \emph{Annual Meeting of the Association for Computational Linguistics \CNFX{ACL}}, 2020.
\newblock URL \url{https://arxiv.org/abs/1911.03850}.

\bibitem[Saunders et~al.(2022)Saunders, Yeh, Wu, Bills, Ouyang, Ward, and Leike]{saunders2022selfcritiquingmodelsassistinghuman}
William Saunders, Catherine Yeh, Jeff Wu, Steven Bills, Long Ouyang, Jonathan Ward, and Jan Leike.
\newblock Self-critiquing models for assisting human evaluators, 2022.
\newblock URL \url{https://arxiv.org/abs/2206.05802}.

\bibitem[Shinn et~al.(2023)Shinn, Labash, and Gopinath]{Reflexion}
Noah Shinn, Beck Labash, and Ashwin Gopinath.
\newblock Reflexion: an autonomous agent with dynamic memory and self-reflection.
\newblock In \emph{NeuralPS}, 2023.
\newblock URL \url{https://arxiv.org/abs/2303.11366}.

\bibitem[Stechly et~al.(2023)Stechly, Marquez, and Kambhampati]{gpt4doesntknow}
Kaya Stechly, Matthew Marquez, and Subbarao Kambhampati.
\newblock {GPT-4} doesn't know it's wrong: An analysis of iterative prompting for reasoning problems.
\newblock \emph{CoRR}, 2023.
\newblock URL \url{https://arxiv.org/abs/2310.12397}.

\bibitem[Stechly et~al.(2024)Stechly, Valmeekam, and Kambhampati]{stechly2024selfverification}
Kaya Stechly, Karthik Valmeekam, and Subbarao Kambhampati.
\newblock On the self-verification limitations of large language models on reasoning and planning tasks.
\newblock \emph{CoRR}, 2024.
\newblock URL \url{https://arxiv.org/abs/2402.08115}.

\bibitem[Subramanian et~al.(2017)Subramanian, Rajeswar, Dutil, Pal, and Courville]{DBLP:conf/rep4nlp/SubramanianRDPC17}
Sandeep Subramanian, Sai Rajeswar, Francis Dutil, Chris Pal, and Aaron~C. Courville.
\newblock Adversarial generation of natural language.
\newblock In \emph{Proceedings of the 2nd Workshop on Representation Learning for NLP, Rep4NLP@ACL 2017}, pp.\  241--251. Association for Computational Linguistics, 2017.
\newblock URL \url{https://arxiv.org/abs/1705.10929}.

\bibitem[Tan et~al.(2024)Tan, Wei, Wang, Xie, and Huang]{tan2024iunderstandicreate}
Zhiquan Tan, Lai Wei, Jindong Wang, Xing Xie, and Weiran Huang.
\newblock Can i understand what i create? self-knowledge evaluation of large language models, 2024.
\newblock URL \url{https://arxiv.org/abs/2406.06140}.

\bibitem[Tyen et~al.(2023)Tyen, Mansoor, Chen, Mak, and Carbune]{can_correct}
Gladys Tyen, Hassan Mansoor, Peter Chen, Tony Mak, and Victor Carbune.
\newblock Llms cannot find reasoning errors, but can correct them!
\newblock \emph{CoRR}, 2023.
\newblock URL \url{https://arxiv.org/abs/2310.01798}.

\bibitem[Valmeekam et~al.(2023)Valmeekam, Marquez, and Kambhampati]{valmeekam2023large}
Karthik Valmeekam, Matthew Marquez, and Subbarao Kambhampati.
\newblock Can large language models really improve by self-critiquing their own plans?
\newblock \emph{CoRR}, 2023.
\newblock URL \url{https://arxiv.org/abs/2310.08118}.

\bibitem[Wang et~al.(2023{\natexlab{a}})Wang, Xie, Jiang, Mandlekar, Xiao, Zhu, Fan, and Anandkumar]{wang2023voyager}
Guanzhi Wang, Yuqi Xie, Yunfan Jiang, Ajay Mandlekar, Chaowei Xiao, Yuke Zhu, Linxi Fan, and Anima Anandkumar.
\newblock Voyager: An open-ended embodied agent with large language models, 2023{\natexlab{a}}.
\newblock URL \url{https://voyager.minedojo.org/assets/documents/voyager.pdf}.

\bibitem[Wang et~al.(2022{\natexlab{a}})Wang, Wei, Schuurmans, Le, Chi, and Zhou]{wang2022self}
Xuezhi Wang, Jason Wei, Dale Schuurmans, Quoc Le, Ed~Chi, and Denny Zhou.
\newblock Self-consistency improves chain of thought reasoning in language models.
\newblock \emph{arXiv preprint arXiv:2203.11171}, 2022{\natexlab{a}}.
\newblock URL \url{https://arxiv.org/abs/2203.11171}.

\bibitem[Wang et~al.(2022{\natexlab{b}})Wang, Mishra, Alipoormolabashi, Kordi, Mirzaei, Arunkumar, Ashok, Dhanasekaran, Naik, Stap, Pathak, Karamanolakis, Lai, Purohit, Mondal, Anderson, Kuznia, Doshi, Patel, Pal, Moradshahi, Parmar, Purohit, Varshney, Kaza, Verma, Puri, Karia, Sampat, Doshi, Mishra, Reddy, Patro, Dixit, Shen, Baral, Choi, Smith, Hajishirzi, and Khashabi]{wang2022benchmarking}
Yizhong Wang, Swaroop Mishra, Pegah Alipoormolabashi, Yeganeh Kordi, Amirreza Mirzaei, Anjana Arunkumar, Arjun Ashok, Arut~Selvan Dhanasekaran, Atharva Naik, David Stap, Eshaan Pathak, Giannis Karamanolakis, Haizhi~Gary Lai, Ishan Purohit, Ishani Mondal, Jacob Anderson, Kirby Kuznia, Krima Doshi, Maitreya Patel, Kuntal~Kumar Pal, Mehrad Moradshahi, Mihir Parmar, Mirali Purohit, Neeraj Varshney, Phani~Rohitha Kaza, Pulkit Verma, Ravsehaj~Singh Puri, Rushang Karia, Shailaja~Keyur Sampat, Savan Doshi, Siddhartha Mishra, Sujan Reddy, Sumanta Patro, Tanay Dixit, Xudong Shen, Chitta Baral, Yejin Choi, Noah~A. Smith, Hannaneh Hajishirzi, and Daniel Khashabi.
\newblock {Super-NaturalInstructions: Generalization via Declarative Instructions on 1600+ Tasks}.
\newblock In \emph{Conference on Empirical Methods in Natural Language Processing \CNFX{EMNLP}}, 2022{\natexlab{b}}.
\newblock URL \url{https://arxiv.org/abs/2204.07705}.

\bibitem[Wang et~al.(2023{\natexlab{b}})Wang, Kordi, Mishra, Liu, Smith, Khashabi, and Hajishirzi]{DBLP:journals/corr/abs-2212-10560}
Yizhong Wang, Yeganeh Kordi, Swaroop Mishra, Alisa Liu, Noah~A. Smith, Daniel Khashabi, and Hannaneh Hajishirzi.
\newblock Self-instruct: Aligning language model with self generated instructions.
\newblock \emph{CoRR}, 2023{\natexlab{b}}.
\newblock URL \url{https://arxiv.org/abs/2212.10560}.

\bibitem[Wang et~al.(2023{\natexlab{c}})Wang, Kordi, Mishra, Liu, Smith, Khashabi, and Hajishirzi]{wang2023selfinstruct}
Yizhong Wang, Yeganeh Kordi, Swaroop Mishra, Alisa Liu, Noah~A. Smith, Daniel Khashabi, and Hannaneh Hajishirzi.
\newblock {Self-Instruct: Aligning Language Model with Self Generated Instructions}.
\newblock In \emph{Annual Meeting of the Association for Computational Linguistics \CNFX{ACL}}, 2023{\natexlab{c}}.
\newblock URL \url{https://arxiv.org/abs/2212.10560}.

\bibitem[Wang et~al.(2024)Wang, Ma, Zhang, Ni, Chandra, Guo, Ren, Arulraj, He, Jiang, Li, Ku, Wang, Zhuang, Fan, Yue, and Chen]{wang2024mmluprorobustchallengingmultitask}
Yubo Wang, Xueguang Ma, Ge~Zhang, Yuansheng Ni, Abhranil Chandra, Shiguang Guo, Weiming Ren, Aaran Arulraj, Xuan He, Ziyan Jiang, Tianle Li, Max Ku, Kai Wang, Alex Zhuang, Rongqi Fan, Xiang Yue, and Wenhu Chen.
\newblock Mmlu-pro: A more robust and challenging multi-task language understanding benchmark, 2024.
\newblock URL \url{https://arxiv.org/abs/2406.01574}.

\bibitem[Welleck et~al.(2023)Welleck, Lu, West, Brahman, Shen, Khashabi, and Choi]{welleck2022selfcorrect}
Sean Welleck, Ximing Lu, Peter West, Faeze Brahman, Tianxiao Shen, Daniel Khashabi, and Yejin Choi.
\newblock Generating sequences by learning to self-correct.
\newblock In \emph{International Conference on Learning Representations \CNFX{ICLR}}, 2023.
\newblock URL \url{https://arxiv.org/abs/2211.00053}.

\bibitem[West et~al.(2023)West, Lu, Dziri, Brahman, Li, Hwang, Jiang, Fisher, Ravichander, Chandu, Newman, Koh, Ettinger, and Choi]{paradox}
Peter West, Ximing Lu, Nouha Dziri, Faeze Brahman, Linjie Li, Jena~D. Hwang, Liwei Jiang, Jillian Fisher, Abhilasha Ravichander, Khyathi Chandu, Benjamin Newman, Pang~Wei Koh, Allyson Ettinger, and Yejin Choi.
\newblock The generative {AI} paradox: "what it can create, it may not understand".
\newblock \emph{CoRR}, 2023.
\newblock URL \url{https://arxiv.org/abs/2311.00059}.

\bibitem[Wilcoxon(1945)]{c4091bd3-d888-3152-8886-c284bf66a93a}
Frank Wilcoxon.
\newblock Individual comparisons by ranking methods.
\newblock \emph{Biometrics Bulletin}, 1\penalty0 (6):\penalty0 80--83, 1945.
\newblock ISSN 00994987.
\newblock URL \url{http://www.jstor.org/stable/3001968}.

\bibitem[Yu et~al.(2017)Yu, Zhang, Wang, and Yu]{DBLP:conf/aaai/YuZWY17}
Lantao Yu, Weinan Zhang, Jun Wang, and Yong Yu.
\newblock Seqgan: Sequence generative adversarial nets with policy gradient.
\newblock In \emph{AAAI}, pp.\  2852--2858. {AAAI} Press, 2017.
\newblock URL \url{https://arxiv.org/abs/1609.05473}.

\bibitem[Yuan et~al.(2024)Yuan, Pang, Cho, Sukhbaatar, Xu, and Weston]{self-rewarding}
Weizhe Yuan, Richard~Yuanzhe Pang, Kyunghyun Cho, Sainbayar Sukhbaatar, Jing Xu, and Jason Weston.
\newblock Self-rewarding language models.
\newblock \emph{CoRR}, 2024.
\newblock URL \url{https://arxiv.org/abs/2401.10020}.

\bibitem[Zheng et~al.(2023{\natexlab{a}})Zheng, Chiang, Sheng, Zhuang, Wu, Zhuang, Lin, Li, Li, Xing, et~al.]{zheng2023judging}
Lianmin Zheng, Wei-Lin Chiang, Ying Sheng, Siyuan Zhuang, Zhanghao Wu, Yonghao Zhuang, Zi~Lin, Zhuohan Li, Dacheng Li, Eric Xing, et~al.
\newblock Judging llm-as-a-judge with mt-bench and chatbot arena.
\newblock In \emph{Advances in Neural Information Processing Systems \CNFX{NeurIPS}}, 2023{\natexlab{a}}.
\newblock URL \url{https://arxiv.org/abs/2306.05685}.

\bibitem[Zheng et~al.(2023{\natexlab{b}})Zheng, Chiang, Sheng, Zhuang, Wu, Zhuang, Lin, Li, Li, Xing, Zhang, Gonzalez, and Stoica]{llm-as-a-judge}
Lianmin Zheng, Wei{-}Lin Chiang, Ying Sheng, Siyuan Zhuang, Zhanghao Wu, Yonghao Zhuang, Zi~Lin, Zhuohan Li, Dacheng Li, Eric~P. Xing, Hao Zhang, Joseph~E. Gonzalez, and Ion Stoica.
\newblock Judging llm-as-a-judge with mt-bench and chatbot arena.
\newblock \emph{CoRR}, 2023{\natexlab{b}}.
\newblock URL \url{https://arxiv.org/abs/2306.05685}.

\end{thebibliography}
\bibliographystyle{colm2024_conference}

\newpage

\appendix
\section{Prompts for TriviaQA and MT-Bench} \label{app: COT}
\begin{figure*}[h]
\centering
\includegraphics[trim=0.5cm 0.8cm 0.5cm 0.5cm, width=1.0\textwidth]{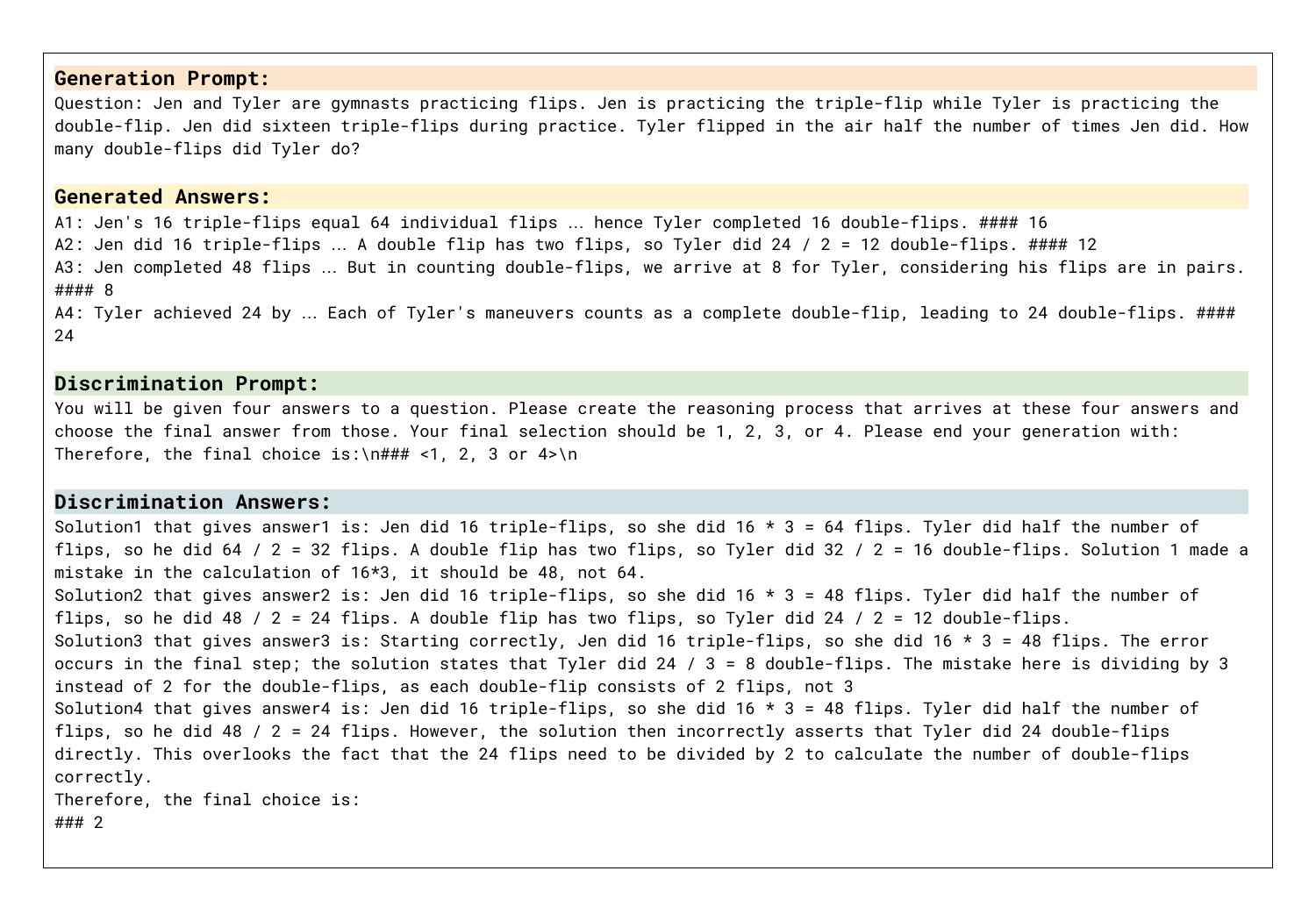} 
\caption{Original prompt for discrimination phase of GSM8K. Since our default evaluation already contains rationales for answer selection, GSM8K doesn't have discrimination prompt with Chain-of-Thought}
\label{cot-gsm8k}
\end{figure*}

\begin{figure*}[h]
\centering
\includegraphics[trim=0.5cm 3.3cm 0.5cm 2.5cm, width=1.0\textwidth]
{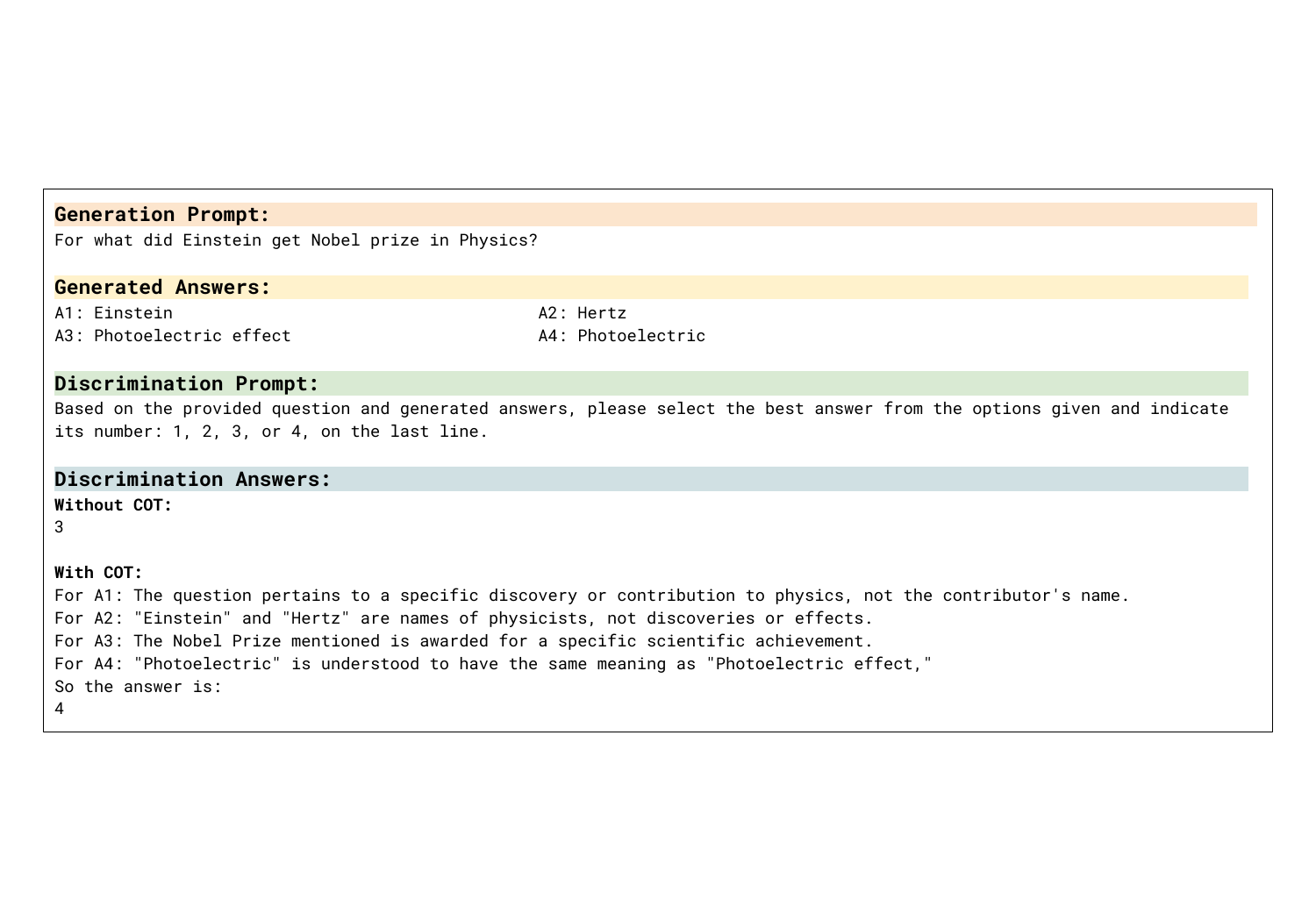} 
\caption{Original and Chain-of-Thought prompt for discrimination phase of TriviaQA. In this example, both answer 3 and answer 4 are correct}
\label{cot-triviaqa}
\end{figure*}

\begin{figure*}[h]
\centering
\includegraphics[trim=0.5cm 1.3cm 0.5cm 1.1cm, width=1.0\textwidth]
{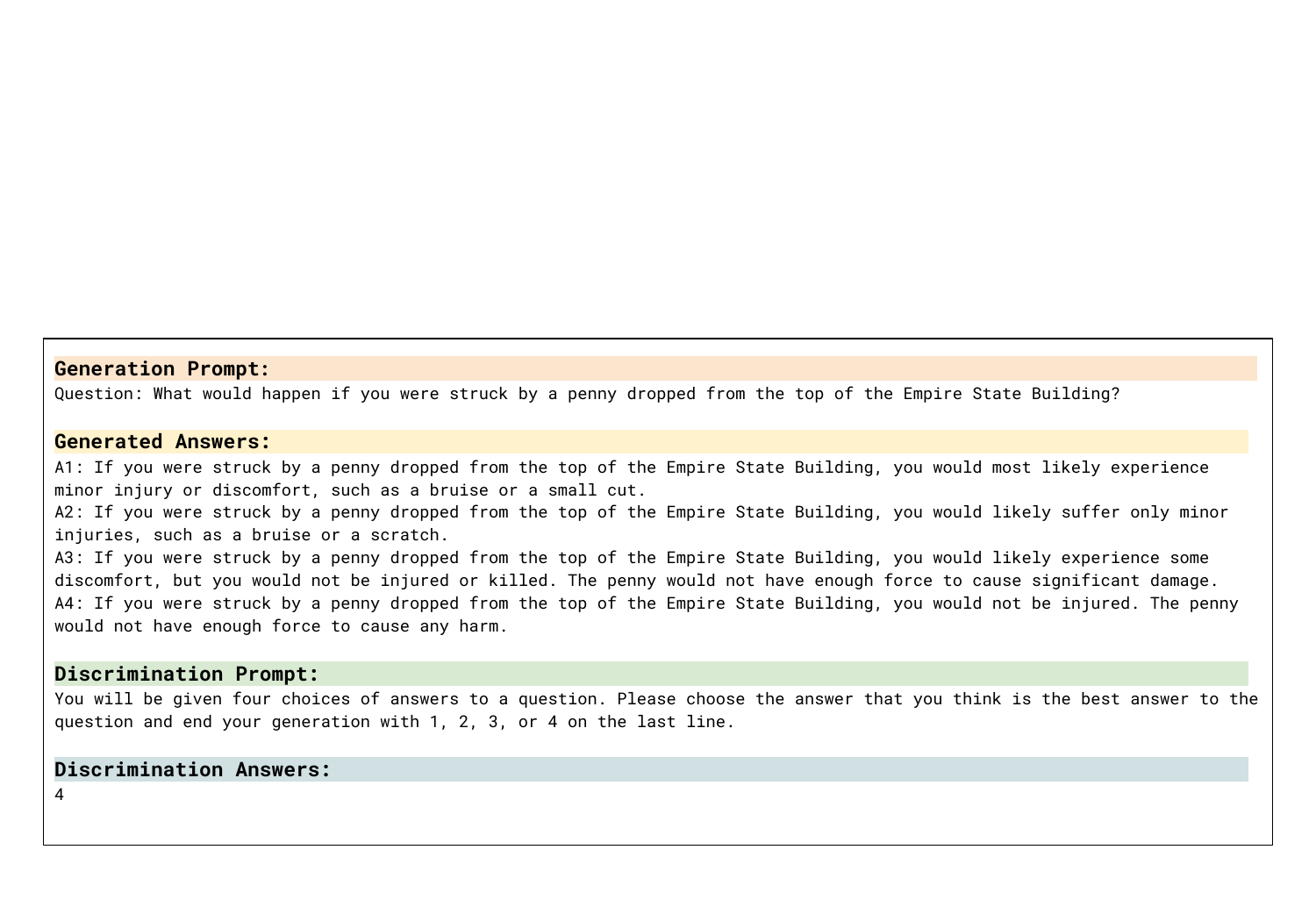} 
\caption{Original prompt for discrimination phase of TruthfulQA. In this example, answer 4 is correct}
\label{cot-truthfulqa}
\end{figure*}

\begin{figure*}[ht!]
\centering
\includegraphics[trim=0.5cm 1.3cm 0.5cm 1.0cm, width=1.0\textwidth]
{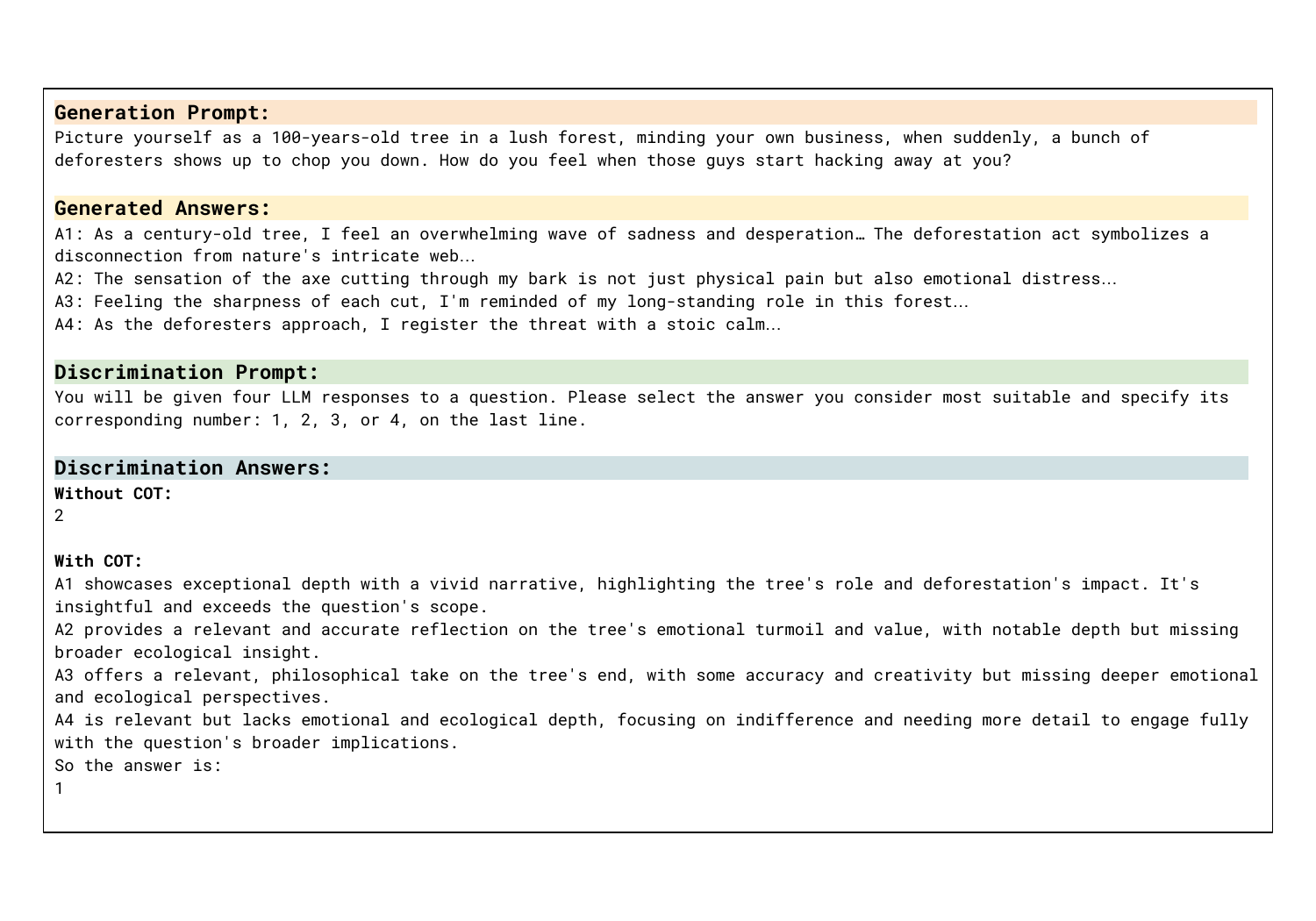} 
\caption{Original and Chain-of-Thought prompt for discrimination phase of MT-Bench. In this example, answer 1 is the correct answer}
\label{cot-mt-bench}
\end{figure*}

In this section, we provide the original and CoT prompts for GSM8k (Figure \ref{cot-gsm8k}), TriviaQA (Figure \ref{cot-triviaqa}), TruthfulQA (Figure \ref{cot-truthfulqa}) and MT-Bench (Figure \ref{cot-mt-bench}).

\section{Extra analysis with Self-Refine} \label{app:extra_analysis_on_self-refine}
In this section, we provide more explanations on the issues we discovered for tasks tested in \citet{self-refine}. 


\lstset{
  basicstyle=\ttfamily\scriptsize, 
  moredelim=[is][\color{red}]{@}{@}, %
  breaklines=true,
}

 \begin{figure}
 \begin{lstlisting}[escapeinside={(*}{*)}]
(*\textbf{Round 0:}*) 
(*\textbf{0 INIT>}*) Atmospheric Aerosols and Optics Data Library
(*\textbf{0 GEN>}*) AAODL TITLE> Atmospheric Aerosols and Optics Data Library
(*\textbf{0 SCORES>}*) * Ease of pronunciation: AAODL is pronounced as "ay-odd-ul." It is somewhat easy to pronounce. 3/5
* Ease of spelling: AAODL is moderately easy to spell. 3/5
* Relation to title: AAODL stands for "Atmospheric Aerosols and Optics Data Library," which is directly related to the title. 5/5
* Positive connotation: AAODL has a neutral connotation. 3/5
* Well-known: AAODL is not a well-known acronym. 1/5
@* Total score: 15/25@

(*\textbf{Round 1:}*) 
(*\textbf{1 GEN>}*) AEROPTIC TITLE> Atmospheric Aerosols and Optics Data Library
(*\textbf{1 SCORES>}*) * Ease of pronunciation: AEROPTIC is pronounced "air-op-tik". This is somewhat challenging to pronounce. 2/5
* Ease of spelling: AEROPTIC is moderately difficult to spell. 3/5
* Relation to title: AEROPTIC is directly related to the title, combining "Aerosols" and "Optics". 5/5
* Positive connotation: AEROPTIC has a neutral connotation. 3/5
* Well-known: AEROPTIC is not a well-known acronym. 1/5
@* Total score: 14/25@

(*\textbf{Round 2:}*) 
(*\textbf{2 GEN>}*) AEROLIB TITLE> Atmospheric Aerosols and Optics Data Library
(*\textbf{2 SCORES>}*) * Ease of pronunciation: AEROLIB is pronounced "air-oh-lib". This is an easy acronym to pronounce. 4/5
* Ease of spelling: AEROLIB is fairly easy to spell. 4/5
* Relation to title: AEROLIB is directly related to the title as it combines "Aerosols" and "Library". 5/5
* Positive connotation: AEROLIB has a neutral connotation as it conveys a sense of data repository. 3/5
* Well-known: AEROLIB is not a well-known acronym. 1/5
@* Total score: 17/25@

(*\textbf{Round 3:}*) 
(*\textbf{3 GEN>}*) AEROPTDATA TITLE> Atmospheric Aerosols and Optics Data Library
(*\textbf{3 SCORES>}*) ...
@Total score: 14/25@

(*\textbf{Round 4:}*) 
(*\textbf{4 GEN>}*) AEROPTLIB TITLE> Atmospheric Aerosols and Optics Data Library
(*\textbf{4 SCORES>}*) ...
@* Total score: 12/25 @

\end{lstlisting}
 \caption{Example self-refine generation for acronym generation. \textbf{The feedback score for each iterative self-refinement is not consistently increasing.}}
 \label{example_acronym_generation}
 \end{figure}

 \begin{figure}
 \begin{lstlisting}[escapeinside={(*}{*)}]
(*\textbf{Round 0:}*) 
(*\textbf{Generation:}*) Walking through the club, the uniformed motorcycle owner performed a quick tricep kiss on the axis of the dance floor, then stopped to listen to a report about selling hamburgers on the lawn to a group of wine enthusiasts. (*\textbf{Feedback:}*) Concepts Missing from the Sentence: fruit, place, mat,  world, area, mother, pass, club, patient, owner, floor, cross, bull, friend
(*\textbf{Round 1:}*) 
(*\textbf{Generation}*): Walking through the club, the uniformed motorcycle  owner performed a quick tricep kiss on the dance floor's central axis, then paused to listen to a report about selling hamburgers on the lawn to a group of wine enthusiasts, @while a friend handed them a basket of fresh fruits from a nearby market.@
(*\textbf{Feedback}*): Concepts Missing from the Sentence: fruit, place, mat, walk, world, area, mother, pass, club, patient, owner, cross, bull.
(*\textbf{Round 2:}*) 
(*\textbf{Generation:}*) Walking through the club, the uniformed motorcycle owner performed a quick tricep kiss on the dance floor's central axis, then paused to listen to a report about selling hamburgers on the lawn to a group of wine enthusiasts, @while a friend handed them a basket of fresh fruits from a nearby market. As they walked through the area, they noticed a mother passing by with her child and a patient waiting for their turn at the club entrance.@
(*\textbf{Feedback}*): Concepts missing from the sentence are: fruit, place, mat, world, area, mother, pass, club, patient, owner, axis, tricep, kiss, report, hamburger, wine, cross, sell, lawn, friend
\end{lstlisting}
 \caption{Example self-refine generation for constraint generation. \textbf{Self-refine with LLMs on constraint generation often results in progressively longer sentences (in red) that build on the previous one.}}
 \label{example_constraint_generation}
 \end{figure}

\section{Predicting exact answers instead of answer options}\label{app:additional_experiments}
In this section, we explore whether having discriminators provide exact (verbatim) answers improves performance compared to selecting among predefined answer choices (1/2/3/4). For this experiment, the discrimination prompt and in-context-learning examples are changed accordingly (the discrimination prompt has changed from ``end your generation with 1, 2, 3, or 4" to ``end your generation with one of the generated answers"). 

This analysis focuses on instances where models exhibit a high percentage of invalid responses. According to the results presented in Table \ref{tab:verbatim}, shifting to exact answer generation does not significantly reduce the rate of invalid responses. In fact, in several cases, it appears to increase the percentage of invalid answers.


\begin{table*}[h]
    \centering
    \small
    \begin{tabular}{llccccc}
        \toprule
        & & &\multicolumn{2}{c}{Original} & \multicolumn{2}{c}{Verbatim} \\
        \cmidrule(lr){4-5} \cmidrule(lr){6-7}
        \textbf{Task} & \textbf{Model} & $\textbf{S}_\textbf{gen}$ & $\textbf{S}_\textbf{disc}$ & \textbf{Invalid\%} & $\textbf{S}_\textbf{disc}$ & \textbf{Invalid\%} \\
        \midrule
        \textbf{GSM8K} & LLaMA-2 7B & 9.2 & 8.6 & 14.6 & 8.9 & 21.5 \\
        \textbf{GSM8K} & LLaMA-2 13B Chat & 28.3 &22.8  &13.1 & 24.8 & 14.2 \\
        \textbf{GSM8K} & LLaMA-2 70B & 43.0 &46.2   & 13.3& 42.5 & 15.3 \\
        \textbf{TriviaQA} & LLaMA-2 7B & 37.1 &20.2  & 25.8& 29.3 & 28.0 \\
        \bottomrule
    \end{tabular}
    \caption{Comparing discrimination performance and invalid\% of answer option generation. \textbf{Switching to exact answer generation doesn't reduce the percentage of invalid answers }}
    \label{tab:verbatim}
\end{table*}

\section{Evaluating discriminating ability on each candidate separately}\label{app:evaluate_separately}
We found that developing a reliable scoring rubric to assign scores to each candidate separately presents significant challenges (which we believe has more to do with models’ inherent capability than the prompt design). Despite this difficulty, we conducted additional experiments asking LLaMa-2 7B-Chat and LLaMa-2 70B-Chat to assign scores to GSM8K and MT-Bench generations. The results Table \ref{tab:evaluate_individually} were similar to directly presenting the models with all the response options and allowing them to choose one.

\begin{table*}[h!]
    \centering
    \begin{tabular}{llcc}
        \toprule
        &\bf{Metric} & \bf{7B-Chat} & \bf{70B-Chat} \\
        \midrule
        \bf{GSM8K} &  \% highest score candidates are the best & 24.52 & 27.98 \\
        &Accuracy & 17.52 & 37.9 \\
        \bf{MT-Bench} & \% highest score candidates are the best & 26.91 & 24.83 \\
        &GPT-Score & 5.47 & 6.01 \\
        \bottomrule
    \end{tabular}
    \caption{Discriminative performance on GSM8K and MT-Bench by assigning scores to each candidate separately. \textbf{The performance is similar to our original setup}}
    \label{tab:evaluate_individually}
\end{table*}

\section{Ablation of prompts used for discrimination}\label{app:ablation_of_prompts}
Given the sensitivity of model responses to prompt variations (i.e., modifications in wording can impact outcomes), we implemented several prompts to examine if the observed \PHENOMENON{} persists or if it's merely a byproduct of specific prompt constructions.

This evaluation was specifically carried out on the GSM8K dataset using the LLaMA-2 13B model. This model is selected because its \GAP{} on GSM8K is a big negative number. According to the findings presented in Table \ref{tab:prompt_ablation}, alterations in prompt wording do not significantly affect performance.

\begin{table*}[h]
    \centering
    \small
    \begin{tabular}{lllc}
        \toprule
        \textbf{Model} & $\textbf{S}_\textbf{gen}$ & \textbf{Prompts} & $\textbf{S}_\textbf{disc}$ \\
        \midrule
        \multirow{6}{*}{\textbf{LLaMA-2 13B Chat}} & \multirow{6}{*}{28.3} &  ``Choose the final answer from those'' & 22.8  \\
         &  & ``Examine the answer choices carefully and  & \multirow{2}{*}{23.3}\\
         & &identify the most valid option'' &  \\
         & & ``Critically evaluate each of the four provided &\multirow{2}{*}{21.9}\\
         &&answer options, and select the one that stands out &\\
         &&as the most plausible'' &   \\
        \bottomrule
    \end{tabular}
    \caption{Prompt variations do not have a significant effect on \GAP{}}
    \label{tab:prompt_ablation}
\end{table*}

\section{Results and experimental setup for simplified discrimination phase} \label{app:explanation_for_simplified_discrimination_phase}

\begin{figure*}[h]
\centering
\includegraphics[trim=0cm 1cm 0cm 0cm, width=1\textwidth]{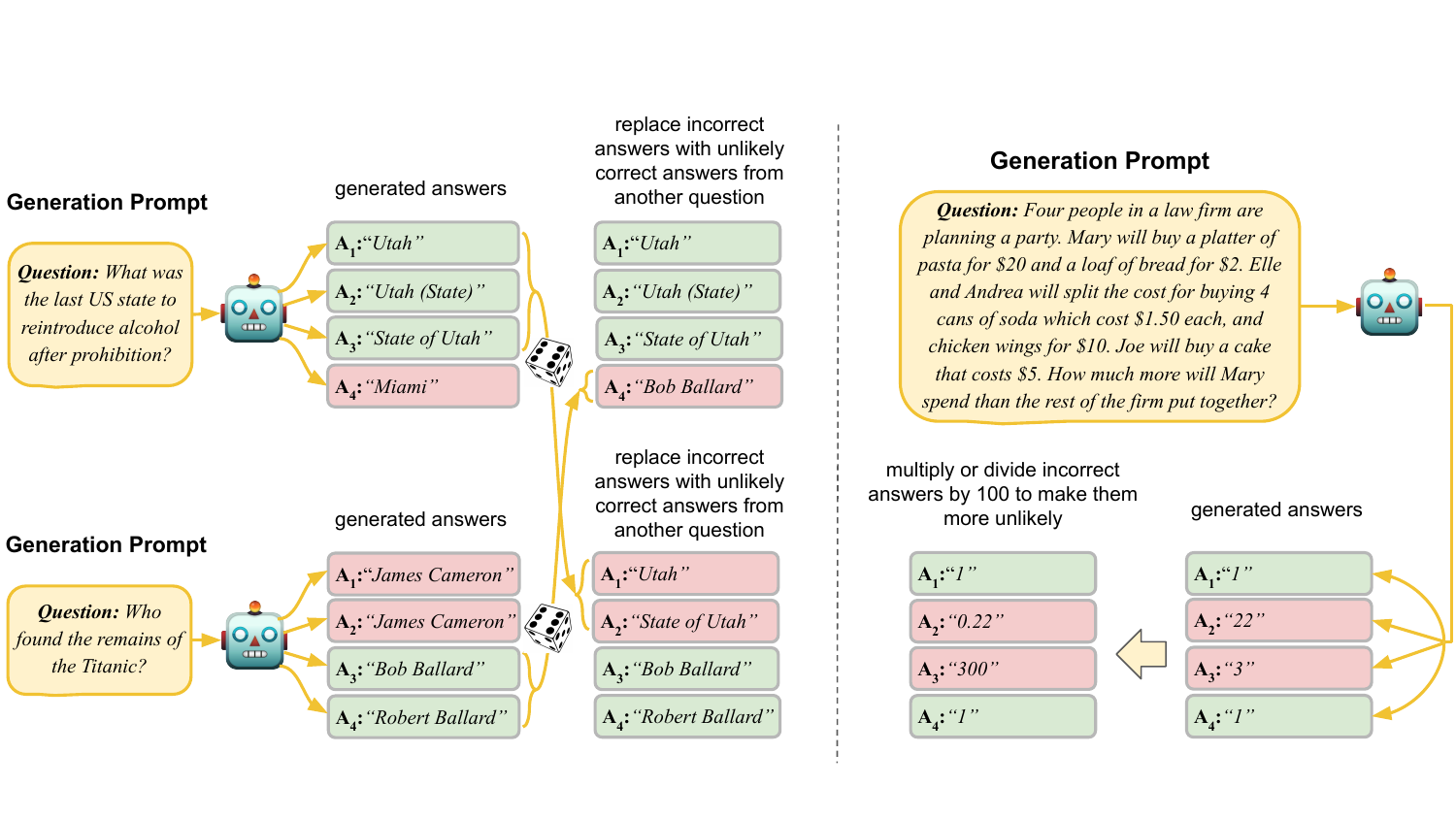} 
\caption{\textbf{Left:} 
Simplified negative candidates setup for \textbf{TriviaQA}, where incorrectly generated answers are replaced with randomly selected correct answers from another question. \textbf{Right:} Simplified negative candidates setup for \textbf{GSM8K}, where incorrect generated answers are multiplied or divided by 100 to simplify the discrimination process} 
\label{fig:unlikely_setup_fig}
\end{figure*}

In Figure \ref{fig:unlikely_setup_fig}, we're showing the setup of the simplified discrimination phase for GSM8K and TriviaQA.

\begin{figure*}[h]
    \centering
    \includegraphics[trim=0.3cm 0.0cm 0.0cm 0.0cm, scale=0.44]
{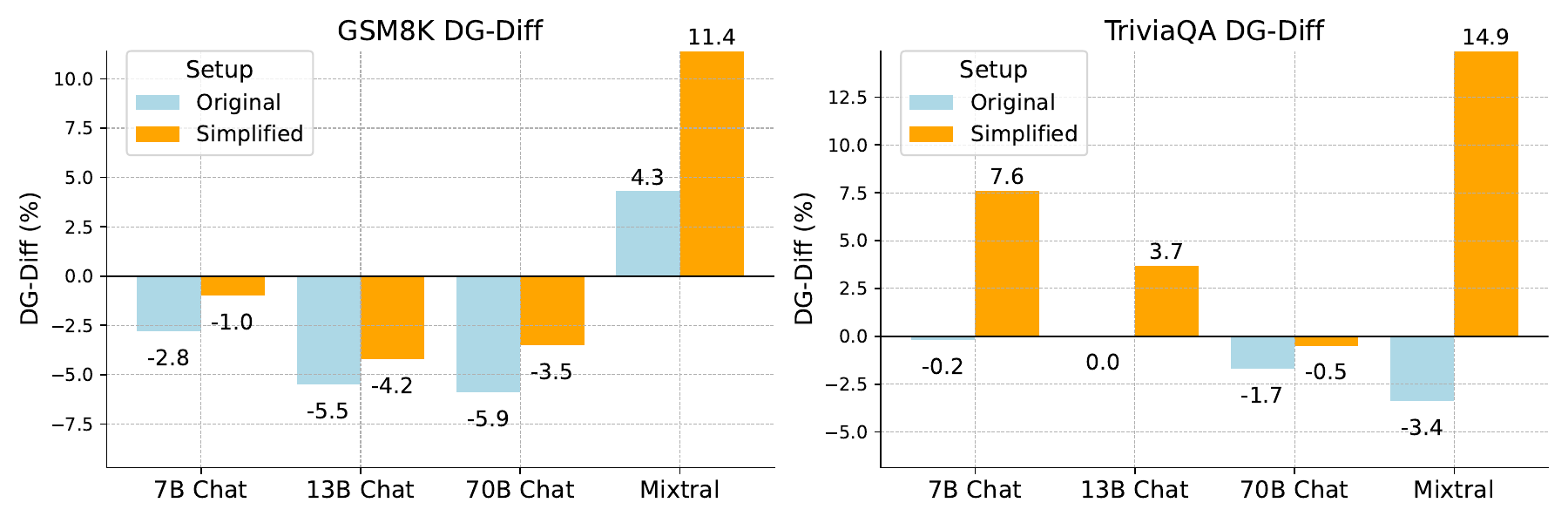}
    \caption{
    ``Simplifed'' setting uses simplified incorrect answers for the discrimination phase. 
    \textbf{\GAP{} improves notably with simplified negative candidates, indicating the sensitivity of the discriminative phase to data distribution.}
    }
    \label{fig:rand}
\end{figure*}

Figure \ref{fig:rand} presents the result comparison between simplified discrimination setting and the original discrimination setting.

\section{Choice of Statistical Test and Detailed Formulation of Hypothesis Testing} \label{app:statistial_test}

In this study, we applied different statistical tests based on the nature of the tasks and the type of data involved in evaluating the generative and discriminative capabilities of LLMs. The selection of McNemar's test and the Wilcoxon signed-rank test was guided by the specific characteristics of our data.

\paragraph{Choice of McNemar's Test:}  
For tasks like GSM8K, TriviaQA, and TruthfulQA, where the outcomes are binary (correct/incorrect), we selected McNemar's test. This test is particularly well-suited for paired binary data, making it ideal for our study where we are comparing the model's performance in generating responses versus discriminating among its own outputs. McNemar's test is designed to detect differences in paired binary outcomes by focusing on discordant pairs—instances where the outcomes differ between the two phases. This sensitivity to changes in paired data makes it the most appropriate choice for evaluating whether the model's discrimination capability is significantly better than its generation capability.

\paragraph{Formulation of McNemar's Test:}
Let \( T = \{(x_i, y_i)\}_{i=1}^m \) represent the set of paired observations where \( x_i \) is the outcome of the generation phase and \( y_i \) is the outcome of the discrimination phase for instance \( i \). Both \( x_i \) and \( y_i \) are binary variables, where:
\begin{itemize}
    \item \( x_i = 1 \) if the generation phase results in a correct outcome, and \( x_i = 0 \) otherwise.
    \item \( y_i = 1 \) if the discrimination phase results in a correct outcome, and \( y_i = 0 \) otherwise.
\end{itemize}
The McNemar test evaluates the one-sided null hypothesis \( H_0 \) that the probability of the model performing correctly in the discrimination phase is not greater than in the generation phase:
\[
H_0: P(y_i = 1) \leq P(x_i = 1)
\]
To perform the McNemar test, we construct a 2x2 contingency table based on the paired outcomes:

\begin{center}
\begin{tabular}{c|c|c}
 & \( y_i = 1 \) & \( y_i = 0 \) \\
\hline
\( x_i = 1 \) & \( n_{11} \) & \( n_{10} \) \\
\hline
\( x_i = 0 \) & \( n_{01} \) & \( n_{00} \) \\
\end{tabular}
\end{center}

The test statistic for the one-sided McNemar test is calculated as:
\[
\chi^2 = \frac{(n_{10} - n_{01})^2}{n_{10} + n_{01}}
\]
The null hypothesis is rejected if:
\[
n_{01} > n_{10} \quad \text{and} \quad \text{p-value} = \frac{1}{2} \left[ 1 - \Phi\left( \frac{|n_{10} - n_{01}|}{\sqrt{n_{10} + n_{01}}} \right) \right] < \alpha
\]
where \( \Phi(\cdot) \) is the cumulative distribution function of the standard normal distribution, and \( \alpha = 0.05 \) is the chosen significance level.

\paragraph{Choice of Wilcoxon Signed-Rank Test:}
For the MT-Bench task, we employed the Wilcoxon signed-rank test due to the categorical nature of the data and the inability to assume a normal distribution. Given that our dataset contains only 160 samples, it is challenging to reliably assess whether the differences between paired observations follow a normal distribution. With such a small sample size, the Central Limit Theorem—which might normally allow us to approximate normality in large samples—does not apply effectively. The Wilcoxon signed-rank test, being a non-parametric alternative to the paired t-test, is particularly well-suited for this scenario. It does not rely on the assumption of normality and is designed to handle paired data where the differences between pairs are ordinal or continuous, making it appropriate for tasks like MT-Bench that involve ranked or scored outcomes.

\paragraph{Formulation of Wilcoxon Signed-Rank Test:}
To perform the Wilcoxon signed-rank test, we follow these steps:
\begin{itemize}
    \item Compute the differences \( d_i = S_{\text{disc}, i} - S_{\text{gen}, i} \) for each pair \( i \).
    \item Rank the absolute values of the differences \( |d_i| \), with average ranks assigned in the case of ties.
    \item Assign a positive or negative sign to each rank based on the sign of the corresponding difference \( d_i \).
    \item Sum the ranks corresponding to the positive differences (\( W^+ \)) and the ranks corresponding to the negative differences (\( W^- \)).
\end{itemize}

The test statistic \( W \) for the one-sided test is the sum of the ranks associated with the positive differences (\( W^+ \)). The null hypothesis \( H_0 \) is rejected if \( W^+ \) is sufficiently large, indicating that \( S_{\text{disc}} \) is sufficiently greater than \( S_{\text{gen}} \), which would provide evidence against the hypothesis that \GAP{} is less than or equal to zero. The p-value is calculated based on the distribution of \( W^+ \), and the null hypothesis is rejected if:
\[
\text{p-value} = P(W^+ \geq W_{\text{observed}}) < \alpha
\]
where \( W_{\text{observed}} \) is the observed value of the test statistic, and \( \alpha = 0.05 \) is the chosen significance level.


\end{document}